\documentclass{article}%
\usepackage{amsmath}
\usepackage{graphicx}%
\usepackage{amsfonts}%
\usepackage{amssymb}
\newtheorem{theorem}{Theorem}

\newtheorem{definition}[theorem]{Definition}
\newtheorem{example}[theorem]{Example}

\newtheorem{lemma}[theorem]{Lemma}

\newtheorem{remark}[theorem]{Remark}

\begin{document}

\title{Estimating Subagging by cross-validation}
\author{Matthieu Cornec, CREST}
\maketitle

\begin{abstract}
\bigskip

In this article, we derive concentration inequalities for the cross-validation
estimate of the generalization error for subagged estimators, both for
classification and regressor. General loss functions and class of predictors
with both finite and infinite VC-dimension are considered. We slightly
generalize the formalism introduced by \cite{DUD03} to cover a large variety
of cross-validation procedures including leave-one-out cross-validation,
$k$-fold cross-validation, hold-out cross-validation (or split sample), and
the leave-$\upsilon$-out cross-validation.

\bigskip

\noindent
An interesting consequence is that the probability upper
bound is bounded by the minimum of a Hoeffding-type bound and a
Vapnik-type bounds, and thus is smaller than 1 even for small
learning set. Finally, we give a simple rule on how to subbag the
predictor. \bigskip


\end{abstract}

\addcontentsline{toc}{section}{Introduction}
\markboth{\uppercase
{Introduction}} {\uppercase{Introduction}} 

\newpage

\section{Introduction and motivation}

\noindent One of the main issue of pattern recognition is to
create a predictor (a regressor or a classifier) which takes
observable inputs in order to predict the unknown nature of an
output. Typical applications range from predicting the figures of
a digitalized zip code to predicting the chance of survival from
clinical measurements. Formally, a predictor $\phi$ is a
measurable map from some measurable space $\mathcal{X}$ to some
measurable space $\mathcal{Y}$. When $\mathcal{Y}$ is a countable
set (respectively $\mathbb{R}^{m}$), the predictor is called a
classifier (respectively a regressor). The strategy of
\textit{Machine Learning }consists in building a learning
algorithm $\Phi$ from both a set of examples and a class of
methods. Typical class of methods are empirical risk minimization
or $k$-nearest neighbors rules. The set of examples consists in
the measurement of $n$ observations $(x_{i},y_{i})_{1\leq i\leq
n}$. Thus, formally, $\Phi$ is a
measurable map from $\mathcal{X}\times\mathcal{\cup}_{n}(\mathcal{X}%
\times\mathcal{Y)}^{n}$ to $\mathcal{Y}$. One of the main issue of
\textit{Statistical Learning }is to analyse the performance of a learning
machine in a probabilistic setting. $(x_{i},y_{i})_{1\leq i\leq n}$ are
supposed to be observations from $n$ independent and identically distributed
(i.i.d.) random variables $(X_{i},Y_{i})_{1\leq i\leq n}\ $with distribution
$\mathbb{P}$. $(X_{i},Y_{i})_{1\leq i\leq n}$ is denoted $\mathcal{D}_{n}$ in
the following and called the learning set. In order to analyse the
performance, it is usual to consider the conditionnal risk of a machine
learning $\Phi$ denoted $\tilde{R}_{n}$, so called the generalization error.
It is defined by the conditional expectation of $L(Y,\Phi(X,\mathcal{D}_{n}))$
given $\mathcal{D}_{n}$ where $(X,Y)\sim\mathbb{P}$ is a random variable
independent of $\mathcal{D}_{n}$, i.e. $\tilde{R}_{n}:=\mathbb{E}%
_{X,Y}(L(Y,\Phi(X,\mathcal{D}_{n}))|\mathcal{D}_{n})$ with $L$ a
cost function from
$\mathcal{Y}^{2}\longrightarrow\mathbb{R}_{+}$. Notice that
$\widetilde{R}_{n}$ is a random variable measurable with respect
to $\mathcal{D}_{n}$.

\bigskip

\noindent Bagging, to be defined formally below, is a procedure building an
estimator by a resample and combine technique. Bagging [bootstrap aggregating]
was introduced by \cite{Breiman96} to reduce the variance of a predictor. From
an original estimator, a bagged regressor is produced by averaging several
replicates trained on bootstrap samples, a bagged classifier is produced by
voting at the majority. It is one of the recent and successful computationally
intensive methods for improving unstable estimation or classification schemes.
It is extremely useful for large, high dimensional data set problems where
finding a good model or classifier in one step is impossible because of the
complexity and scale of the problem. Regarding prediction error, the method
often compares favorably with the original predictor, and also, in situations
with substantial noise, with other ensemble methods such as boosting or
randomization. Hence it is very important to understand the reasons for its
successes, and also for its occasional failures. However, even if it has
attracted much attention and is frequently applied, important questions remain
unanswered theoretically. In this article, we study a variant of bagging
called Subagging [Subsample aggregating] that has appeared in
\cite{Friedman00} and \cite{Buhlman00}. It is more accessible for analysis and
has also substantial computational advantages. The subagged estimator will be
denoted by $\Phi^{B}(X,\mathcal{D}_{n})$ or $\Phi_{n}^{B}(X)$ in the following.

\bigskip

\noindent Important questions are: \textit{Is the generalization error of a
subagged predictor lower than the original predictor, i.e }$\widetilde{R}%
_{n}(\Phi_{n}^{B})\leq\widetilde{R}_{n}(\Phi)$\textit{? The distribution
$\mathbb{P}$ of the generating process being unknown, can we estimate the
generalization error of a subagged predictor? }Our strategy is the following:
after briefly emphasizing the difficulty to provide a general answer to the
first question, we will concentrate on the second question. To estimate the
generalization error of a subagged predictor, we propose to use an adapted
cross-validation estimator denoted by $\widehat{R}_{CV}^{Out}(\Phi)$.

\bigskip

\noindent\cite{Breiman96} aggregates regression trees to build
random forest and calls this process bagging. \cite{Buja02} prove
that the bagged functional is always smooth in some sense.
\cite{Buja00} also show that bagging can increase both bias and
variance. \cite{Friedman00} prove that (in the limit of infinite
samples) bagging reduces the variance of non-linear components of
the Taylor decomposition while leaving the linear part unaffected.
\cite{Buhlman00} consider non-differentiable and discontinuous
predictors and concentrate on the asymptotic smoothing effect of
bagging on neighborhood of discontinuities of decision surfaces.
\cite{Grandvalet04} brings new argument to explain bagging
effect: bagging's improvement/deteriations are explained by the
goodness/badness of highly influential examples.
\cite{Elisseeff04} prove the effect of bagging on the stability
of a learning method and derive non asymptotic bounds for the
approximation error of the bagging predictor. An interesting
asymptotic result was derived in \cite{BIA08} : asymptotically,
bagging of weak predictors can produce a strong learner, namely
the bayes classifier. However, a
general answer to the following non-asymptotic question $\widetilde{R}_{n}(\Phi_{n}^{B}%
)\leq\widetilde{R}_{n}(\Phi)?$ seems hard to reach in a  general framework.
Using Gauss-Markov theorem, \cite{Grandvalet04} shows that both bagged and
unbagged predictor are unbiased, thus the variance of the unbagged predictor
is lower than the variance of the bagged one. \cite{Buja00} exhibit general
quadratic statistics for which the bagged predictor increase both variance and
bias. Thus, we propose to estimate directly the generalization error of the
subagged predictor by an adapted cross-validation procedure.
The latter is inspired by  \cite{PET07}, who proposed to use the
left-out example of the bootstrap samples.

\bigskip

\noindent In the general setting, the cross-validation procedures
include leave-one-out cross-validation, $k$-fold
cross-validation, hold-out cross-validation (or split sample),
leave-$\upsilon$-out cross-validation (or Monte Carlo
cross-validation or bootstrap cross-validation). With the
exception of \cite{BUR89}, theoretical investigations of multifold
cross-validation procedures have first concentrated on linear
models (\cite{Li87} ;\cite{SHAO93} ; \cite{ZHA93}). Results of
\cite{DGL96} and \cite{GYO02} are discussed in Section 3. The
first finite sample results are due to Wagner and Devroye
\cite{DEWA79} and concern $k$-local rules algorithms under
leave-one-out and hold-out cross-validation. More recently,
\cite{HOL96, HOL96bis} derived finite sample results for
$\upsilon$-out cross-validation, $k-$fold cross-validation, and
leave-one-out cross-validation for ERM over a class of predictors
with finite VC-dimension in the realisable case (the
generalization error is equal to zero). \cite{BKL99} have
emphasized when $k-$fold can beat $\upsilon$-out cross-validation
in the particular case of $k$-fold predictor. \cite{KR99} has
extended such results in the case of stable algorithms for the
leave-one-out cross-validation procedure. \cite{KEA95} also
derived results for hold-out cross-validation for ERM, but their
arguments rely on the traditional notion of VC-dimension. In the
particular case of ERM over a class of predictors with finite
VC-dimension but with general cross-validation procedures,
\cite{COR09A} derived probability upper bounds. \cite{COR09B}
derived upper bounds for general cross-validation estimate of the
generalization error of stable predictors that do no make
reference to VC-dimension. However, these bounds obtained are
called ''sanity check bounds''\ since they are not better than
classical Vapnik-Chernovenkis's bounds.\bigskip

\noindent We introduce our \textbf{main result} for symmetric
cross-validation procedures (i.e. the probability for an
observation to be in the test set is independent of its index) in
the special case of empirical risk minimization (ERM). We divide
the learning sample into two samples: the training sample and the
test sample, to be defined below. We denote by $p_{n}$ the
percentage of elements in the test sample. Suppose that
$\mathcal{H}$ holds, to be defined below. Suppose also that
$\phi_{n}$ is an empirical risk minimizer. Then, we have for all
$\varepsilon>0$,
\[
\Pr(\widetilde{R}_{n}(\Phi_{n}^{B})-\hat{R}_{CV}^{Out}\geq\varepsilon)\leq
\min(B_{ERM}(n,p_{n},\varepsilon),V_{ERM}(n,p_{n},\varepsilon))<1,
\]

with

\begin{itemize}
\item $B_{ERM}(n,p_{n},\varepsilon)=\displaystyle     \min((2np_{n}%
+1)^{4V_{\mathcal{C}}/p_{n}}\exp(-n\varepsilon^{2}),(2n(1-p_{n}%
)+1)^{^{\frac{4V_{\mathcal{C}}}{1-p_{n}}}}\exp(-n\varepsilon^{2}/9))$

\item $V_{ERM}(n,p_{n},\varepsilon)=\displaystyle     \exp(-2np_{n}%
\varepsilon^{2})$.\bigskip
\end{itemize}

\noindent The term $B(n,p_{n},\varepsilon)$ is a
Vapnik-Chernovenkis-type bound controlled by the size of the
training sample $n(1-p_{n})$ whereas the term
$V(n,p_{n},\varepsilon)$ is the minimum between a Hoeffding-type
term controlled by the size of the test sample $np_{n}$, a
polynomial term controlled by the size of the training sample.
%and a
%Vapnik-like exponential term.
This bound can be interpreted as a quantitative answer to a
trade-off issue. As the percentage of observations in the test
sample $p_{n}$ increases, the term $V(n,p_{n},\varepsilon)$
decreases but the term $B(n,p_{n},\varepsilon)$ increases. Other
similar bounds are derived for infinite VC-dimension machine
learning in the stability framework.

%Notice that this bound is worse than the Vapnik-like bound and
%thus can be called a ''sanity-check bound'' in the spirit of
%\cite{KR99}.

\noindent The main interest of the previous results is in the
following

\begin{itemize}
\item our bounds are valid for machine learning with both finite and infinite
VC-dimension. In the latter, it is sufficient that the machine learning
satisfies some stablity property as introduced in chapter 2. As a
motivation, we quote the following list of algorithms satisfying stability
properties: regularization networks, ERM, k-nearest rules, boosting.

\item our bounds are strictly less than $1$ for any size of learning set. Thus
it is also valid for small samples.
\end{itemize}

\bigskip

\noindent Using these probability bounds, we can then deduce that the
expectation of the difference between the generalization error and the
cross-validation estimate
\[
\mathbb{E}_{\mathcal{D}_{n}}\widetilde{R}_{n}(\Phi_{n}^{B})-\hat{R}_{CV}%
^{Out}\leq\min(\sqrt{1/np_{n}},6\sqrt{\frac{V_{\mathcal{C}}(\ln(n(1-p_{n}%
))+2)}{n(1-p_{n})}}).
\]
\noindent Eventually, we define a splitting rule on how to chose the
percentage of elements $p_{n}^{\star}$ in the test sample in order to get both
a low generalization error together with a good approximation rate. We
derive for this optimal choice of $p^{\star}$ a bound of the form

$$\Pr(\widetilde{R}_{n}(\Phi_{n}^{B,\star})-\hat{R}_{CV}^{Out}(p_{n}^{\star
})\geq\varepsilon)=O_{n}((n+1)^{8V_{\mathcal{C}}}\exp(-2n(\varepsilon
-2\sqrt{2}V_{\mathcal{C}}^{1/2}\sqrt{\ln(n)/n})^{2}/(1-\exp(-2\varepsilon
^{2})).$$

\bigskip

\noindent The paper is organized as follows. We detail the main
cross-validation procedures and we summarize the previous results
for the estimation of generalization error. In Section 3, we
introduce the main notations and definitions. Finally, in Section
4, we introduce our results, in terms of concentration
inequalities.

\section{Main notations}

\noindent In the following, we follow the notations of cross-validation
introduced in \cite{COR09A}.

\noindent We will consider the following shorter notations inspired by the
literature on empirical processes. In the sequel, we will denote
$\mathcal{Z}:=\mathcal{X\times Y}$, and $(Z_{i})_{1\leq i\leq n}%
:=((X_{i},Y_{i}))_{1\leq i\leq n}$ \ the learning set. For a given loss
function $L$ and a given class of predictors $\mathcal{G}$, we define a new
class $\mathcal{F}$ of functions from $\mathcal{Z}$ to $\mathbb{R}_{+}$ by
$\mathcal{F}:=\{\psi\in\mathbb{R}_{+}^{\mathcal{Z}}|\psi(Z)=L(Y,\phi
(X)),\phi\in\mathcal{G}\}$. For a machine learning $\Phi$, we have the natural
definition $\Psi(Z,\mathcal{D}_{n})=L(Y,\Phi(X,\mathcal{D}_{n})).$ With these
notations, the conditional risk $\widetilde{R}_{n}\ $is the expectation of
$\Psi(Z,\mathcal{D}_{n})$ with respect to $\mathbb{P}$ conditionally on
$\mathcal{D}_{n}$: $\widetilde{R}_{n}:=\mathbb{E}_{Z}[\Psi(Z,\mathcal{D}%
_{n})\mid\mathcal{D}_{n}]$ with $Z\sim\mathbb{P}$ independent of
$\mathcal{D}_{n}$. In the following, if there is no ambiguity, we will also
allow the following notation $\psi(X,\mathcal{D}_{n})$ instead of
$\Psi(X,\mathcal{D}_{n})$.

\noindent To define the accurate type of cross-validation procedure, we
introduce binary vectors. Let $V_{n}=(V_{n,i})_{1\leq i\leq n}$ be a vector of
size $n$. $V_{n}$ is a binary vector if for all $1\leq i\leq n,V_{n,i}%
\in\{0,1\}$ and if $\sum_{i=1}^{n}V_{n,i}\neq0$. Consequently, we
can define the subsample associated with it:
$\mathcal{D}_{V_{n}}:=\{Z_{i}\in \mathcal{D}_{n}|V_{n,i}=1,1\leq
i\leq n\}$. We define a weighted empirical measure on
$\mathcal{Z}$
\[
\mathbb{P}_{n,V_{n}}:=\frac{1}{\sum_{i=1}^{n}V_{n,i}}\sum_{i=1}^{n}%
V_{n,i}\delta_{Z_{i}},%
\]
with $\delta_{Z_{i}}$ the Dirac measure at $\{Z_{i}\}$. We also define a
weighted empirical error $\mathbb{P}_{n,V_{n}}\psi$ where $\mathbb{P}%
_{n,V_{n}}\psi$ stands for the usual notation of the expectation
of $\psi$ with respect to $\mathbb{P}_{n,V_{n}}$. For
$\mathbb{P}_{n,1_{n}}$, with $1_{n}$ the binary vector of size
$n$ with $1$ at every coordinate, we will use the traditional
notation $\mathbb{P}_{n}$. For a predictor trained on a
subsample, we define
\[
\psi_{V_{n}}(.):=\Psi(.,\mathcal{D}_{V_{n}}).
\]

%\begin{definition}[Weighted empirical measure] For a given non-negative vector
%$(V_{n,i}%
%)_{1 \leq i \leq n}%
%$ of size $n$, we define a weighted empirical measure:
%$$ \mathbb{P}_{n,V_n}
%:= \frac{1}{\sum_i V_{n,i}}
%\sum_{i=1}^{n} V_{n,i} \delta_{(X_{i},Y_{i}%
%)}$$
%with $\delta_{(X_{i},Y_{i})}$ the Dirac measure at $(X_{i}%
%,Y_{i})$
%\end{definition}

\noindent With the previous notations, notice that the predictor trained on
the learning set $\psi(.,\mathcal{D}_{n})\ $can be denoted by $\psi_{1_{n}%
}(.)$. We will allow the simpler notation $\psi_{n}(.)$. The learning set is
divided into two disjoint sets: the training set of size $n(1-p_{n})$ and the
test set of size $np_{n}$, where $p_{n}$ is the percentage of elements in the
test set. To represent the training set, we define $V_{n}^{tr}$ a random
binary vector of size $n$ independent of $\mathcal{D}_{n}$. $V_{n}^{tr}$ is
called the training vector. We define the test vector by $V_{n}^{ts}%
:=1_{n}-V_{n}^{tr}$ to represent the test set.

\bigskip

\noindent The distribution of $V_{n}^{tr}$ characterizes all the subagging
procedures described in the previous section. Using our notations, we can now
define the bagged predictor.

\bigskip

\begin{definition}
[Subagged regressor]The subagged predictor build from $\phi_{n}$
denoted
$\phi_{n}^{B}$ is defined by:%
\[
\phi_{n}^{B}(.):=\mathbb{E}_{V_{n}^{tr}}\phi_{V_{n}^{tr}}(.).
\]
\end{definition}

\noindent In the case of classifiers, the bagging rule corresponds to the vote
by majority. We suppose in this case that $\mathcal{Y}=\{1,\ldots,M\}$.

\begin{definition}
[Subagged classifier]Cross-validated subagged classifiers of
$\phi_{n}^{B}$
defined by:%
$$
\phi_{n}^{B}(X)    :=\arg\min_{k\in\{1,\ldots,M\}}\mathbb{E}_{V_{n}^{tr}%
}L(k,\Phi(X,\mathcal{D}_{V_{n}^{tr}}))
$$
\end{definition}

\noindent We can now define the cross-validation estimator.

\begin{definition}
[Cross-validated subagged estimator]Cross-validated subagged estimates of
$\phi_{n}^{B}$ denoted can be defined in two different ways by:%
\[
\widehat{R}_{CV}^{Out}(\Phi_{n}^{B}):=\mathbb{E}_{V_{n}^{tr}}\mathbb{P}%
_{n,V_{n}^{ts}}(\psi_{V_{n}^{tr}})
\]

and%
\[
\widehat{R}_{CV}^{In}(\Phi_{n}^{B}):=\mathbb{E}_{V_{n}^{tr}}\mathbb{P}%
_{n,V_{n}^{tr}}(\psi_{V_{n}^{tr}})
\]
\end{definition}

\begin{remark}
Recall that $\mathbb{E}_{V_{n}^{tr}}\mathbb{P}_{n,V_{n}^{ts}}(\psi_{V_{n}%
^{tr}})$ is the conditional expectation of $\mathbb{P}_{n,V_{n}^{ts}}%
(\psi_{V_{n}^{tr}})$ with respect to the random vector $V_{n}^{tr}$ given
$\mathcal{D}_{n}$.
\end{remark}

\begin{remark}
The cross-validated subagged estimate differs from the usual cross-validation
estimate of $\hat{R}_{CV}^{Out}(\psi_{n}^{B})$ which is equal to
$\mathbb{E}_{U_{n}^{tr}}\mathbb{P}_{n,U_{n}^{ts}}(\psi_{U_{n}^{tr}}^{B})$ with
$U_{n}^{tr}$ the training vector as defined in chapter 1.
\end{remark}

\bigskip

\noindent We will give here a few examples of distributions of $V_{n}^{tr}$ to
show we retrieve subagging procedures described previously. Suppose $n/k$ is
an integer. The $k$-fold subagging procedure divides the data into $k$ equally
sized folds. It then produces a predictor by training on $k-1$ folds. This is
repeated for each fold, and the trained predictors are averaged to form the
subagged predictor.

\begin{example}
[$k$-fold cross-validation]%
\begin{align*}
\Pr(V_{n}^{tr}  &  =(\underbrace{0,\ldots,0}_{n/k\text{ observations}%
},\underbrace{1,\ldots,1}_{n(1-1/k)\text{ observations}}))=\frac{1}{k}\\
\Pr(V_{n}^{tr}  &  =(\underbrace{1,\ldots,1}_{n/k\text{ observations}%
},\underbrace{0,\ldots,0}_{n/k\text{ observations}},\underbrace{1,\ldots
,1}_{n(1-2/k)\text{ observations}}))=\frac{1}{k}\\
&  \ldots\\
\Pr(V_{n}^{tr}  &  =(\underbrace{1,\ldots,1}_{n(1-1/k)\text{ observations}%
},\underbrace{0,\ldots,0}_{n/k\text{ observations}}))=\frac{1}{k}.%
\end{align*}
\end{example}

\noindent We provide another popular example: the leave-one-out
cross-validation. In leave-one-out cross-validation, a single sample of size
$n$ is used. Each member of the sample in turn is removed, the full modeling
method is applied to the remaining $n-1$ members, and the fitted model is
applied to the hold-backmember.

\begin{example}
[leave-one-out cross-validation]%
\begin{align*}
\Pr(V_{n}^{tr}  &  =(0,1,\ldots,1))=\frac{1}{n}\\
\Pr(V_{n}^{tr}  &  =(1,0,1,\ldots,1))=\frac{1}{n}\\
&  \ldots\\
\Pr(V_{n}^{tr}  &  =(1,\ldots,1,0))=\frac{1}{n}.%
\end{align*}
\end{example}

\section{Results for the cross-validated subagged regressor}

\subsection{VC Framework}

\subsubsection{Notations and definition}

\noindent We denote by $R_{opt}$ the minimal generalization error attained
among the class of predictors $\mathcal{C}$, $R_{opt}=\inf_{\phi\in
\mathcal{C}}R(\phi)$. In the sequel, we suppose that $\phi_{n}$ belongs to
some $\mathcal{C}$. Notice that $R_{opt}$ is a parameter of the unknown
distribution $\mathbb{P}_{(X,Y)}$ whereas $\widetilde{R}_{n}$ is a random variable.

\bigskip

\noindent At last, recall the definitions of:

\begin{definition}
[Shatter coefficients]Let $\mathcal{A}$ be a collection of measurable sets.
For $(z_{1,\ldots,}z_{n})$ $\in\{\mathbb{R}^{d}\}^{n}$, let $N_{\mathcal{A}%
}(z_{1,\ldots,}z_{n})$ be the number of differents sets in
\[
\{\{z_{1},\ldots,z_{n}\}\cap A;A\in\mathcal{A}\}
\]

The n-shatter coefficient of $\mathcal{A}$ is
\[
\mathcal{S}(\mathcal{A},n)=\max_{(z_{1,\ldots,}z_{n})\in\{\mathbb{R}^{d}%
\}^{n}}N_{\mathcal{A}}(z_{1,\ldots,}z_{n})
\]

That is, the shatter coefficient is the maximal number of different subsets of
$n$ points that can be picked out by the class of sets $\mathcal{A}$.
\end{definition}

\noindent and

\begin{definition}
[VC dimension]Let $\mathcal{A}$ be a collection of sets with $\mathcal{A}%
\geq2.$ The largest integer $k\geq1$ for which $\mathcal{S}(\mathcal{A}%
$,$k)=2^{k}$ is denoted by $V_{\mathcal{C}}$, and it is called the
Vapnik-Chernovenkis dimension (or VC dimension) of the class $\mathcal{A}$. If
$\mathcal{S}(\mathcal{A}$,$n)=2^{n}$ for all n, then by definition
$V_{\mathcal{C}}=\infty.$
\end{definition}

\noindent A class of predictors $\mathcal{C}$ is said to have a
finite VC-dimension $V_{\mathcal{C}}$ if the dimension of the
collection of sets
$\{A_{\phi,t}:\phi\in\mathcal{C},t\in\lbrack0,1]\}$ is equal to
$V_{\mathcal{C}}$, where
$A_{\phi,t}=\{(x,y)/L(y,\phi(x))>t\}$.\bigskip

\subsubsection{Results}

\noindent In the sequel, we suppose that the cross-validation is
symmetric (i.e. $\Pr(V_{n,i}=1)$ is independent of $i$) and the
number of elements in the training set is constant and equal to
$np_{n}$, that the training sample and the test sample are
disjoint and that the number of observations in the training
sample and in the test sample are respectively $n(1-p_{n})$ and
$np_{n}$. Moreover, we suppose also that $\phi_{n}$ belongs to a
class of predictor with finite VC-dimension. Suppose also that
$L$ is bounded in the following way: $L(Y,\phi(X))\leq
C(h(Y,\phi(X))$ with $C$ convex function -bounded itself by $1$
on the support of $h(Y,\phi_{V_n^{tr}}(X))$ for simplicity-, and
$h$ such that for any $0<\lambda<1$, we have
$h(y,\lambda\phi(x_{1})+(1-\lambda)\phi (x_{2})\leq \lambda
h(y,\phi(x_{1})+(1-\lambda)h(y,\phi(x_{2})$. We will also suppose
that the predictors are symmetric according to the training
sample, i.e. the predictor does not depend on the order of the
observations in $\mathcal{D}_{n}$. \textbf{We denote these
hypotheses by $\mathcal{H}$.}

\bigskip

\begin{remark}
Typical upperbounding convex cost functions are : the hinge loss
$C(x)=(1+x)_{+}$, the exponential loss $C(x)=e^{x}$, the logit
loss $C(x)=\log_{2}(1+e^{x})$.
\end{remark}

\bigskip

\noindent
%$\mathbb{P}$ stands for $\mathbb{P}_{%
%\mathcal{D}_{n}}$.
We will show upper bounds of the kind $\Pr(\widetilde{R}_{n}(\Phi_{n}%
^{B})-\hat{R}_{CV}^{Out}\geq\varepsilon)\leq\min(B(n,p_{n},\varepsilon
),V(n,p_{n},\varepsilon))$ with $\varepsilon>0$. The term $B(n,p_{n}%
,\varepsilon)$ is a Vapnik-Chernovenkis-type bound whereas the term $V(n,p_{n}%
,\varepsilon)$ is a Hoeffding-type term controlled by the size of the test
sample $np_{n}$. This bound can be interpreted as a quantitative answer
to a trade-off question. As the percentage of observations in the test sample
$p_{n}$ increases, the $V(n,p_{n},\varepsilon)$ term decreases but the
$B(n,p_{n},\varepsilon)$ term increases.

\bigskip

\begin{theorem}
[Absolute error for symmetric cross-validation]\label{thm:sym}

\noindent Suppose that $\mathcal{H}$ holds. Then, we have for all
$\varepsilon>0$,
$$\Pr(\widetilde{R}_{n}(\Phi_{n}^{B})-\hat{R}_{CV}^{Out}\geq\varepsilon)
\leq
\min(B_{sym}(n,p_{n},\varepsilon),V_{sym}(n,p_{n},\varepsilon))<1
$$
with

\begin{itemize}
\item $B_{sym}(n,p_{n},\varepsilon)=\displaystyle     (2np_{n}%
+1)^{4V_{\mathcal{C}}/p_{n}}e^{-n\varepsilon^{2}}$

\item $V_{sym}(n,p_{n},\varepsilon)=\displaystyle     \exp(-2np_{n}%
\varepsilon^{2}).$
\end{itemize}
\end{theorem}

\bigskip

\begin{remark}
We do not require $\phi_{n}$ to be an empirical risk minimizer.
\end{remark}

\noindent{\bfseries Proof.}

\bigskip

\noindent We have $\widetilde{R}_{n}(\Phi_{n}^{B})=\mathbb{P}\psi_{n}%
^{B}=\mathbb{P}L(Y,\mathbb{E}_{V_{n}^{tr}}\phi_{V_{n}^{tr}}(X))$.
Since $C$ is a convex function -bounded itself by $1$ on the
support of $h(Y,\phi_{V_n^{tr}}(X))$-, and $h$ linear in the
second variable, we get
\[
\widetilde{R}_{n}(\Phi_{n}^{B})\leq\mathbb{P}C(h(Y,\mathbb{E}_{V_{n}^{tr}%
}\phi_{V_{n}^{tr}}(X))\leq\mathbb{E}_{V_{n}^{tr}}\mathbb{P}C(h(Y,\phi
_{V_{n}^{tr}}(X))
\]

\noindent Then, we split according to $\mathbb{E}_{V_{n}^{tr}}\mathbb{P}%
_{n,V_{n}^{ts}}C(h(Y,\phi_{V_{n}^{tr}}(X))$:
\begin{align*}
\widetilde{R}_{n}(\Phi_{n}^{B})  &  \leq\mathbb{E}_{V_{n}^{tr}}\mathbb{P}%
_{n,V_{n}^{ts}}C(h(Y,\phi_{V_{n}^{tr}}(X))+\mathbb{E}_{V_{n}^{tr}%
}(\mathbb{P-\mathbb{P}}_{n,V_{n}^{ts}})C(h(Y,\phi_{V_{n}^{tr}}(X))\\
&  =\hat{R}_{CV}^{Out}+\mathbb{E}_{V_{n}^{tr}}(\mathbb{P-\mathbb{P}}%
_{n,V_{n}^{ts}}\mathbb{)}C(h(Y,\phi_{V_{n}^{tr}}(X)
\end{align*}

\noindent Thus, we obtain: $\Pr(\widetilde{R}_{n}(\psi_{n}^{B})-\hat{R}%
_{CV}^{Out}\geq\varepsilon)\leq\Pr(\mathbb{E}_{V_{n}^{tr}}(\mathbb{P-}%
\mathbb{P}_{n,V_{n}^{ts}}\mathbb{)}C(h(Y,\phi_{V_{n}^{tr}}(X)\geq
\varepsilon)$.

\bigskip

\noindent To prove our result, we proceed now in two steps. For this, we
consider
$$\mathbb{E}_{V_{n}^{tr}}(\mathbb{P}_{n,V_{n}^{ts}}C(h(Y,\phi
_{V_{n}^{tr}}(X))-\mathbb{P}C(h(Y,\phi_{V_{n}^{tr}}(X))\mathbb{)}$$
in two different ways

\begin{enumerate}
\item using conditional Hoeffding's inequality,

\item using Vapnik-Chernovenkis-type inequality to bound the supremum over a class.
\end{enumerate}

\bigskip

\begin{enumerate}
\item First, by conditional Hoeffding arguments (for a proof, see e.g. chapter 1),
$$\Pr(\widetilde{R}_{n}(\Phi_{n}^{B})-\hat{R}_{CV}^{Out}\geq\varepsilon
)\leq\exp(-2np_{n}\varepsilon^{2}).$$

\item Secondly, we derive the bound:%
\begin{align*}
\Pr(\widetilde{R}_{n}(\Phi_{n}^{B})-\hat{R}_{CV}^{Out} \geq\varepsilon)  &
\leq\Pr(\mathbb{E}_{V_{n}^{tr}}(\mathbb{P-\mathbb{P}}_{n,V_{n}^{ts}}%
\mathbb{)}C(h(Y,\phi_{V_{n}^{tr}}(X))\geq\varepsilon)\\
&  \leq\Pr(\mathbb{E}_{V_{n}^{tr}}\sup_{\phi \in \mathcal{C}}(\mathbb{P-\mathbb{P}}%
_{n,V_{n}^{ts}}\mathbb{)}C(h(Y,\phi(X))\geq\varepsilon).
\end{align*}
\end{enumerate}

\noindent Recall a useful lemma (for the proof, see Appendices).

\begin{lemma}
Under the assumptions $\mathcal{H}$, we have for all, $\varepsilon>0,$%

\[
\Pr\mathbb{(E}_{V_{n}^{tr}}\sup_{\phi\in\mathcal{C}}(\mathbb{P-}%
\mathbb{P}_{n,V_{n}^{tr}}\mathbb{)}C(h(Y,\phi(X))\geq\varepsilon
)\leq(\mathcal{S}(2np_{n},\mathcal{C}))^{4/p_{n}}e^{-n\varepsilon^{2}}.%
\]
\bigskip
\end{lemma}

\noindent and we also have (for the proof, see e.g. \cite{DGL96}): $\forall
n,\mathcal{S}(n,\mathcal{C})\leq(n+1)^{V_{\mathcal{C}}}$.

\bigskip

\noindent Thus, it follows that $\Pr(\widetilde{R}_{n}(\Phi_{n}^{B})-\hat
{R}_{CV}^{Out}\geq\varepsilon)\leq(2np_{n}+1)^{4V_{\mathcal{C}}/p_{n}%
}e^{-n\varepsilon^{2}}$.

\noindent Putting altogether, we get $\Pr(\widetilde{R}_{n}(\Phi_{n}^{B}%
)-\hat{R}_{CV}^{Out}\geq\varepsilon)\leq\min(\exp(-2np_{n}\varepsilon
^{2}),(2np_{n}+1)^{4V_{\mathcal{C}}/p_{n}}e^{-n\varepsilon^{2}}).$

\bigskip

$\Box$

\begin{theorem}
[Absolute error for symmetric cross-validation]Suppose that $\mathcal{H}$
holds. Then, we have for all $\varepsilon>0$,
\[
\Pr(\widetilde{R}_{n}(\Phi_{n}^{B})-\hat{R}_{CV}^{In}\geq\varepsilon)\leq
\min(B_{sym}(n,p_{n},\varepsilon),V_{sym}(n,p_{n},\varepsilon))<1
\]

with

\begin{itemize}
\item $B_{sym}(n,p_{n},\varepsilon)=  (2n(1-p_{n})%
+1)^{\frac{4V_{\mathcal{C}}}{1-p_{n}}}e^{-n\varepsilon^{2}}$

\item $V_{sym}(n,p_{n},\varepsilon)=    \exp(-2np_{n}%
\varepsilon^{2}).$
\end{itemize}
\end{theorem}

\noindent{\bfseries      Proof. }

\bigskip

\noindent We proceed as previously: $\widetilde{R}_{n}(\Phi_{n}^{B}%
)=\mathbb{P}\Phi_{n}^{B}=\mathbb{P}L(Y,\mathbb{E}_{V_{n}^{tr}}\phi
_{V_{n}^{tr}}(X))\leq\mathbb{P}C(h(Y,\mathbb{E}_{V_{n}^{tr}}\phi
_{V_{n}^{tr}}(X))\leq\mathbb{E}_{V_{n}^{tr}}\mathbb{P}C(h(Y,\phi
_{V_{n}^{tr}}(X)$.

\noindent We then split this quantity according to $\mathbb{E}_{V_{n}^{tr}}\mathbb{P}%
_{n,V_{n}^{tr}}C(h(Y,\phi_{V_{n}^{tr}}(X)$%

\begin{align*}
\widetilde{R}_{n}(\Phi_{n}^{B})  &  \leq\mathbb{E}_{V_{n}^{tr}}\mathbb{P}%
_{n,V_{n}^{tr}}C(h(Y,\phi_{V_{n}^{tr}}(X)+\mathbb{E}_{V_{n}^{tr}%
}(\mathbb{P-\mathbb{P}}_{n,V_{n}^{ts}}\mathbb{)}C(h(Y,\phi_{V_{n}^{tr}%
}(X)\\
&  =\hat{R}_{CV}^{In}+\mathbb{E}_{V_{n}^{tr}}(\mathbb{P-\mathbb{P}}%
_{n,V_{n}^{tr}}\mathbb{)}C(h(Y,\phi_{V_{n}^{tr}}(X)).
\end{align*}

\noindent Thus, we get%

\begin{align*}
\Pr(\widetilde{R}_{n}(\Phi_{n}^{B})-\hat{R}_{CV}^{In}  &  \geq\varepsilon
)\leq\Pr(\mathbb{E}_{V_{n}^{tr}}(\mathbb{P-}\mathbb{P}_{n,V_{n}^{tr}%
}\mathbb{)}C(h(Y,\phi_{V_{n}^{tr}}(X))\geq\varepsilon)\\
&  \leq\Pr(\mathbb{E}_{V_{n}^{tr}}\sup_{\phi\in\mathcal{C}}(\mathbb{P-}%
\mathbb{P}_{n,V_{n}^{tr}}\mathbb{)}C(h(Y,\phi(X))\geq\varepsilon).
\end{align*}

\noindent Recall two useful results (for the proof, see e.g. chapter 1)

\begin{lemma}
\label{monlemme}Under the assumptions $\mathcal{H}$, we have for all
$\varepsilon>0,$%

\[
\Pr\mathbb{(E}_{V_{n}^{tr}}\sup_{\phi\in\mathcal{C}}(\mathbb{P}(\phi
)-\mathbb{P}_{n,V_{n}^{tr}}(\phi))\geq\varepsilon)\leq(\mathcal{S}%
(2n(1-p_{n}),\mathcal{C}))^{4/(1-p_{n})}e^{-n(1-p_{n})\varepsilon^{2}}.%
\]
\end{lemma}

$\bigskip$

$\Box$

\noindent In the special case of empirical risk minimization, we can obtain a
stronger result.

\begin{theorem}
[Absolute error for symmetric cross-validation]\noindent Suppose that
$\mathcal{H}$ holds. Suppose also that $\phi_{n}$ is based on empirical risk
minimization. But instead of minimizing $\widehat{R}_{n}(\phi)$, we suppose
$\phi_{n}$ minimizes $\frac{1}{n}\sum_{i=1}^{n}C(h(Y_{i},\phi(X_{i}))$. For
simplicity, we suppose the infimum is attained i.e. $\phi_{n}=\arg\min
_{\phi\in\mathcal{C}}\frac{1}{n}\sum_{i=1}^{n}C(h(Y_{i},\phi(X_{i}))$. Then,
we have for all $\varepsilon>0$,
\[
\Pr(\widetilde{R}_{n}(\Phi_{n}^{B})-\hat{R}_{CV}^{Out}\geq\varepsilon)\leq
\min(B_{ERM}(n,p_{n},\varepsilon),V_{ERM}(n,p_{n},\varepsilon))<1,
\]

with

\begin{itemize}
\item $B_{ERM}(n,p_{n},\varepsilon)=\displaystyle     \min((2np_{n}%
+1)^{4V_{\mathcal{C}}/p_{n}}\exp(-n\varepsilon^{2}),(2n(1-p_{n}%
)+1)^{^{\frac{4V_{\mathcal{C}}}{1-p_{n}}}}\exp(-n\varepsilon^{2}/9))$

\item $V_{ERM}(n,p_{n},\varepsilon)=\displaystyle     \exp(-2np_{n}%
\varepsilon^{2}).$
\end{itemize}
\end{theorem}

\bigskip

\begin{remark}
\begin{enumerate}
\item The assumption $\phi_{n}=\arg\min_{\phi\in\mathcal{C}}\frac{1}{n}%
\sum_{i=1}^{n}C(h(Y_{i},\phi(X_{i}))$ is not so restrictive, since in practice
in order to numerically minimizes $\frac{1}{n}\sum_{i=1}^{n}L(Y_{i},\phi
(X_{i}))$, one looks for $C$ convex such that for all $x,y,$ $L(y,\phi(x))\leq
C(h(y,\phi(x))$.

\item Thanks to the Hoeffding's part, the bound is always smaller than $1$, so
it remains valid for small samples. For bigger samples, we will
prefer the Vapnik-Chernovenkis's part.
\end{enumerate}
\end{remark}

\noindent{\bfseries      Proof. }

\noindent Appying the previous result, we have $\Pr(\widetilde{R}_{n}(\Phi
_{n}^{B})-\hat{R}_{CV}^{Out}\geq\varepsilon)\leq\min(\exp(-2np_{n}%
\varepsilon^{2}),(2np_{n}+1)^{4V_{\mathcal{C}}/p_{n}}\exp(-n\varepsilon^{2}))$.

\bigskip

\noindent Recall that $\widetilde{R}_{n}(\Phi_{n}^{B})-\hat{R}_{CV}^{Out}%
\leq\mathbb{E}_{V_{n}^{tr}}(\mathbb{P}C(h(Y,\phi_{V_{n}^{tr}}(X))-\mathbb{P}%
_{n,V_{n}^{ts}}C(h(Y,\phi_{V_{n}^{tr}}(X))\mathbb{)}$.

\bigskip

\noindent We need the following lemma (for a proof, see chapter 1):
$\mathbb{E}_{V_{n}^{tr}}\mathbb{P}_{n,V_{n}^{ts}}C(h(Y,\phi_{V_{n}^{tr}%
}(X))\geq\mathbb{P}_{n}C(h(Y,\phi_{n}(X))$ since $\phi_{n}=\arg\min_{\phi
\in\mathcal{C}}\frac{1}{n}\sum_{i=1}^{n}C(h(Y_{i},\phi(X_{i}))$.

\bigskip

\noindent Denote $\psi(Z):=C(h(Y,\phi(X)))$ with $Z:=(X,Y)$. We have the
following natural notation $\psi_{V_{n}^{tr}}(Z):=C(h(Y,\phi_{V_{n}^{tr}%
}(X)))$.

\bigskip

\noindent We thus get
\[
\Pr(\widetilde{R}_{n}(\Phi_{n}^{B})-\hat{R}_{CV}^{Out}\geq3\varepsilon)\leq
\Pr(\mathbb{E}_{V_{n}^{tr}}(\mathbb{P}\psi_{V_{n}^{tr}}\mathbb{-P}%
_{n,V_{n}^{ts}}\psi_{V_{n}^{tr}})\geq3\varepsilon)\leq\Pr(\mathbb{E}%
_{V_{n}^{tr}}(\mathbb{P}\psi_{V_{n}^{tr}}\mathbb{-P}_{n}\psi_{n}%
)\geq3\varepsilon)
\]
and by splitting according to $\mathbb{\mathbb{\mathbb{\mathbb{P}}}}\psi
_{opt}$, we have:
\begin{align*}
\Pr(\widetilde{R}_{n}(\Phi_{n}^{B})-\hat{R}_{CV}^{Out}\geq3\varepsilon)  &
\leq\Pr(\mathbb{E}_{V_{n}^{tr}}(\mathbb{P}\psi_{V_{n}^{tr}}\mathbb{-\mathbb{P}%
}_{n,V_{n}^{tr}}\psi_{V_{n}^{tr}}\mathbb{+\mathbb{\mathbb{P}}}_{n,V_{n}^{tr}%
}\psi_{V_{n}^{tr}}-\mathbb{\mathbb{\mathbb{\mathbb{P}}}}\psi_{opt}%
\mathbb{+\mathbb{\mathbb{P}}}\psi_{opt}\mathbb{-P}_{n}\psi_{n})\geq
3\varepsilon)\\
&  \leq\Pr(\mathbb{E}_{V_{n}^{tr}}\sup_{\psi\in\mathcal{F}}(\mathbb{P}%
\psi\mathbb{-\mathbb{P}}_{n,V_{n}^{tr}}\psi)\geq\varepsilon)+\Pr(\sup_{\psi
\in\mathcal{F}}(\mathbb{\mathbb{P}}_{n,V_{n}^{tr}}\psi\mathbb{-P}\psi
)\geq\varepsilon)\\
&  \quad+\Pr(\sup_{\psi\in\mathcal{F}}(\mathbb{\mathbb{P}}\psi\mathbb{-P}%
_{n}\psi)\geq\varepsilon).
\end{align*}

\bigskip

\noindent Recall the following lemma (for the proof, see e.g.chapter 1),

\begin{lemma}
Under the assumption of Proposition \ref{largets1}, we have for all
$\varepsilon>0,$%
\[
\Pr\mathbb{(E}_{V_{n}^{tr}}\sup_{\psi\in\mathcal{F}}(\mathbb{\mathbb{P}%
}_{n,V_{n}^{tr}}\psi\mathbb{-P}\psi)\geq\varepsilon)\leq(\mathcal{S}%
(2n(1-p_{n}),\mathcal{C}))^{\frac{4}{1-p_{n}}}e^{-n\varepsilon^{2}}%
\]
\noindent and symmetrically%

\[
\Pr\mathbb{(E}_{V_{n}^{tr}}\sup_{\psi\in\mathcal{F}}(\mathbb{P}\psi
\mathbb{-\mathbb{P}}_{n,V_{n}^{tr}}\psi)\geq\varepsilon)\leq(\mathcal{S}%
(2n(1-p_{n}),\mathcal{C}))^{\frac{4}{1-p_{n}}}e^{-n\varepsilon^{2}}.%
\]
\end{lemma}

\noindent Then, we get
\begin{align*}
\Pr(\widetilde{R}_{n}(\psi_{n}^{B})-\hat{R}%
_{CV}^{Out}\geq3\varepsilon)&\leq2(\mathcal{S}(2n(1-p_{n}),\mathcal{C}%
))^{\frac{4}{1-p_{n}}}e^{-n\varepsilon^{2}}+(\mathcal{S}(2n,\mathcal{C}%
))^{4}e^{-n\varepsilon^{2}} \\
& \leq3(2n(1-p_{n})+1)^{^{\frac{4V_{\mathcal{C}}%
}{1-p_{n}}}}e^{-n\varepsilon^{2}}.
\end{align*}

\bigskip

\noindent This implies in turn that
\[
\Pr(\widetilde{R}_{n}(\psi_{n}^{B})-\hat{R}_{CV}^{Out}\geq\varepsilon
)\leq(2n(1-p_{n})+1)^{^{\frac{4V_{\mathcal{C}}}{1-p_{n}}}}\exp(-n\varepsilon
^{2}/9).
\]

\bigskip

\noindent Putting altogether, we get
\begin{align*}
\Pr(\widetilde{R}_{n}(\psi_{n}%
^{B})-\hat{R}_{CV}^{Out}\geq\varepsilon)& \leq\min(\exp(-2np_{n}\varepsilon
^{2}),(2np_{n}+1)^{4V_{\mathcal{C}}/p_{n}}e^{-n\varepsilon^{2}},\\
& \quad \quad (2n(1-p_{n}%
)+1)^{^{\frac{4V_{\mathcal{C}}}{1-p_{n}}}}\exp(-n\varepsilon^{2}/9))
\end{align*}

\bigskip

$\Box$

\bigskip

\begin{theorem}
\noindent Suppose that $\mathcal{H}$ holds. Suppose also and that $n/k$ is an
integer. Then, we have also for all $\varepsilon>0$,%
\[
\Pr(\widetilde{R}_{n}(\psi_{n}^{B})-\hat{R}_{CV}^{Out}\geq\varepsilon)\leq
\min(B_{k}(n,p_{n},\varepsilon),V_{k}(n,p_{n},\varepsilon))
\]

with

\begin{itemize}
\item $B_{k}(n,p_{n},\varepsilon)=\displaystyle     (2n/k+1)^{4kV_{\mathcal{C}%
}}\exp(-n\varepsilon^{2})$

\item {\small $V_{k}(n,p_{n},\varepsilon)=\displaystyle     \min\left(
\exp(-2n/k\varepsilon^{2}),2^{\frac{1}{p_{n}}}\exp\left(  -\frac{n\epsilon
^{2}}{64(\sqrt{V_{\mathcal{C}}\ln(2(2n/k+1))}+2)}\right)  \right)  .$}
\end{itemize}
\end{theorem}

\noindent{\bfseries      Proof.}

\noindent The proofs starts as previously. We have
$$\Pr(\hat{R}_{CV}^{Out}%
-\widetilde{R}_{n}(\psi_{n}^{B})\geq\varepsilon)\leq\Pr(\mathbb{E}_{V_{n}%
^{tr}}(\mathbb{P}_{n,V_{n}^{ts}}\psi_{V_{n}^{tr}}\mathbb{-P}\psi_{V_{n}^{tr}%
})\geq\varepsilon)\leq\exp(-2np_{n}\varepsilon^{2})$$

\noindent but we also have
\begin{align*}
\Pr(\hat{R}_{CV}^{Out}-\widetilde{R}_{n}(\psi_{n}^{B})\geq\varepsilon)&\leq
\Pr(\mathbb{E}_{V_{n}^{tr}}(\sup_{\psi\in\mathcal{F}}(\mathbb{P}_{n,V_{n}%
^{ts}}\psi\mathbb{-P}\psi)\geq\varepsilon) \\
& \leq 2^{\frac{1}{p_{n}}}\exp\left(
-\frac{n\epsilon^{2}}{64(\sqrt{V_{\mathcal{C}}\ln(2(2np_{n}+1))}+2)}\right).
\end{align*}
\noindent according to chapter 1.

$\Box$

\bigskip

\noindent Following the previous results, we can obtain results for the
expectation of the difference $\widetilde{R}_{n}(\psi_{n}^{B})-\hat{R}%
_{CV}^{Out}$

\begin{theorem}
[$L_{1}$ error]\noindent Suppose that $\mathcal{H}$ holds. Suppose also and
that $n/k$ is an integer. Then, we have also for all $\varepsilon>0$,%
\[
\mathbb{E}_{\mathcal{D}_{n}}\left(  \widetilde{R}_{n}(\psi_{n}^{B})-\hat
{R}_{CV}^{Out}\right)  \leq\sqrt{1/np_{n}}%
\]

\noindent Furthermore, suppose also that $\phi_{n}$ is based on empirical risk
minimization. But instead of minimizing $\widehat{R}_{n}(\phi)$, we suppose
$\phi_{n}$ minimizes $\frac{1}{n}\sum_{i=1}^{n}C(h(Y_{i},\phi(X_{i}))$. For
simplicity, we suppose the infimum is attained i.e. $\phi_{n}=\arg\min
_{\phi\in\mathcal{C}}\frac{1}{n}\sum_{i=1}^{n}C(h(Y_{i},\phi(X_{i}))$. Then,
we have,%

\[
\mathbb{E}_{\mathcal{D}_{n}}\left(  \widetilde{R}_{n}(\psi_{n}^{B})-\hat
{R}_{CV}^{Out}\right)  \leq\min(\sqrt{1/np_{n}},6\sqrt{\frac{V_{\mathcal{C}%
}(\ln(n(1-p_{n}))+2)}{n(1-p_{n})}})
\]
\end{theorem}

{\bfseries      Proof. }

\noindent We just need to apply the previous results together with the
following useful lemma (for a proof, see e.g.\cite{DGL96}):

\begin{lemma}
\label{Monlemme}\label{lem esperance} Let $X$ be a nonnegative random
variable. Let $K,C$ nonnegative real such that $C\geq1$. Suppose that for all
$\varepsilon>0$, $\mathbb{P}(X\geq\varepsilon)\leq C\exp(-K\varepsilon^{2})$.
Then, we have
\[
\mathbb{E}X\leq\sqrt{\frac{\ln(C)+2}{K}}.
\]
\end{lemma}

$\bigskip$

$\Box$

\subsection{Stability framework}

\subsubsection{Introduction to stability}

\noindent To avoid the traditional analysis in the VC framework, notions of
stability have been intensively worked through in the late 90's \cite{KEA95},
\cite{BE01}, \cite{BE02}, \cite{KUT02}, and \cite{KUNIY02}. The object of
stability framework is the learning algorithm rather than the space of
classifiers. The learning algorithm is a map (effective procedure) from data
sets to classifiers. An algorithm is stable at a learning set $\mathcal{D}%
_{n}$ if changing one point in $\mathcal{D}_{n}$ yields only a small change in
the output hypothesis. Several different notions of algorithmic stability are
described. The attraction of such an approach is that it avoids the
traditional notion of VC-dimension, and allows to focus on a wider class of
learning algorithms than empirical risk minimization. For example, this
approach provides generalization error bounds for regularization-based
learning algorithms that have been difficult to analyze within the VC
framework such as boosting. If a map is stable, exponential bounds on
generalization error may be obtained. As a motivation, we quote the following
list of algorithms satisfying stability properties: regularization networks,
ERM, k-nearest rules, boosting.

\subsubsection{Definitions and notations of stability}

\noindent The basic idea is that an algorithm is stable at a
training set $\mathcal{D}_{n}$ if changing one point in
$\mathcal{D}_{n}$ yields only a small change in the output
hypothesis. Formally, a learning algorithm maps a weighted
training set into a predictor space. Thus, stability can be
translated into a Lipschitz condition for this mapping with high
probability.

\noindent To be more formal, following \cite{COR09B}, we define a distance
between two weighted empirical errors:

\begin{definition}
[Total variation]Let $\mathbb{P}_{n,V_{n}}$ and $\mathbb{P}_{n,U_{n}}$ be two
empirical measures on $\mathcal{Z}$ with respect to the binary vectors $V_{n}$
and $U_{n}$. We do not assume their support to be equal. The distance between
them is defined as their total variation:%
\[
||\mathbb{P}_{n,U_{n}}-\mathbb{P}_{n,V_{n}}||=\sup_{A\in\mathcal{P}%
(\mathcal{Z})}|(\mathbb{P}_{n,U_{n}}-\mathbb{P}_{n,V_{n}})(A)|.
\]
\end{definition}

\begin{example}
In the case of leave-one-out (i.e. $\sum_{i=1}^{n}U_{n,i}=n-1$), we have:
\[
||\mathbb{P}_{n,U_{n}}-\mathbb{P}_{n}||=\frac{2}{n}.%
\]

In the case of leave-$\nu$-out, we get:%
\[
||\mathbb{P}_{n,U_{n}}-\mathbb{P}_{n}||=\frac{2\nu}{n}.%
\]
\end{example}

\noindent At least, we need a distance $d$ on the set $\mathcal{F}$. Let us
quote three important examples. Let $\psi_{1},\psi_{2}$ $\in\mathcal{F}$. The
uniform distance is defined by: $d_{\infty}(\psi_{1},\psi_{2})=\sup
_{Z\in\mathcal{Z}}|\psi_{1}(Z)-\psi_{2}(Z)|$, the $L_{1}$-distance by:
$d_{1}(\psi_{1},\psi_{2})=\mathbb{P}|\psi_{1}-\psi_{2}|$ , the error-distance
$d_{e}(\psi_{1},\psi_{2})=|\mathbb{P(}\psi_{1}-\psi_{2})|$. It is important to
notice that what matters here is not an absolute distance between the original
class of predictors $\mathcal{G}$ seen as functions but the distance with the
respect to the loss or/and the distribution $\mathbb{P}$. In particular, for
the $L_{1}$-distance, we do not care about the behavior of the original
predictors $\phi_{1}$ and $\phi_{2}$ outside the support of $\mathbb{P}%
$. At last, notice that we always have $d_{e}\leq d_{1}\leq d_{\infty}$.

\bigskip

\noindent We are now in position to define the different notions
of stability of a learning algorithm which cover notions
introduced by \cite{KUNIY02}. We begin with the notion of weak
stability. In essence, it says that for any given resampling
vectors, the distance between two predictors is controlled with
high probability by the distance between the resampling vectors.
As a motivation, notice that algorithms such as Adaboost
(\cite{KUNIY02}) satisfies this property. With the previous
notations, we have:

\begin{definition}
[Weak stability]Let $\mathcal{D}_{n}=(Z_{i})_{1\leq i\leq n}$ be a learning
set. Let $\lambda,(\delta_{n,p_{n}})_{n,p_{n}}$ be nonnegative real numbers. A
learning algorithm $\Psi$ is said to be weak $(\lambda,(\delta_{n,p_{n}%
})_{n,p_{n}},d)$ stable if for any training vector $U_{n}$ whose sum is equal
to $n(1-p_{n})$:%
\[
\Pr(d(\psi_{U_{n}},\psi_{n})\geq\lambda||\mathbb{P}_{n,U_{n}}-\mathbb{P}%
_{n}||)\leq\delta_{n,p_{n}}.%
\]
\end{definition}

\noindent Notice that in the former definition $\Pr$ stands for $\mathbb{P}%
^{\otimes n}$. Indeed, $\psi_{n}$ is trained with $n$ observations, drawn
independently from $\mathbb{P}$. A stronger notion is to consider $\psi_{n}$
trained with $n-1$ observations drawn independently from $\mathbb{P}$ and an
additionnal general observation $z$. We consider the stronger notion of strong
stability. As a motivation, notice that algorithms such as Empirical Risk
Minimization with finite VC\ dimension (\cite{KUNIY02}) satisfies this property.

\begin{definition}
[Strong stability]Let $z\in\mathcal{Z}$. Let $\mathcal{D}_{n}=\mathcal{D}%
_{n-1}\cup\{z\}$ be a learning set. Let
$\lambda,(\delta_{n,p_{n}})_{n,p_{n}}$ be nonnegative real
numbers. A learning algorithm $\Psi$ is said to be strong
$(\lambda,(\delta_{n,p_{n}})_{n,p_n},d)$ stable if for any
training vector $U_{n}$ whose
sum is equal to $n(1-p_{n})$:%
\[
\Pr(d(\psi_{U_{n}},\psi_{n})\geq\lambda||\mathbb{P}_{n,U_{n}}-\mathbb{P}%
_{n}||)\leq\delta_{n,p_{n}}.%
\]
\end{definition}

\noindent What we have in mind for classical algorithms is $\delta_{n,p_{n}%
}=O_{n}(p_{n}\exp(-n(1-p_{n}))$. We can state the last definition in other
words. Let $V_{n}^{tr}$ be a training vector with distribution $\mathbb{Q}$
such that the number of elements in the training set is constant and equal to
$n(1-p_{n})$. Notice then that the former definition also implies that
$\sup_{U_{n}\in\text{support}(\mathbb{Q)}}\mathbb{P}(\frac{d(\psi_{U_{n}}%
,\psi_{n})}{||\mathbb{P}_{n,U_{n}}-\mathbb{P}_{n}||}\geq\lambda)\leq
\delta_{n,p_{n}}$, where support$(\mathbb{Q)}$ stands for the support of
$\mathbb{Q}$. The previous notion stands for any $U_{n}$ having the same
support of $\mathbb{Q}$. A stronger hypothesis would be that the previous
probability stands uniformly over $U_{n}$ in support$(\mathbb{Q)}$. This leads
formally to the notion of cross-validation stability. To be more accurate:

\begin{definition}
[Cross-validation weak stability]Let
$\mathcal{D}_{n}=(Z_{i})_{1\leq i\leq n}$ a learning set. Let
$V_{n}^{tr}$ a training vector with distribution $\mathbb{Q}$.
Let $\lambda,(\delta_{n,p_{n}})_{n,p_{n}}$ be nonnegative real
numbers. A learning algorithm $\Psi$ is said to be weak
$(\lambda,(\delta _{n,p_{n}})_{n,p_{n}},d,\mathbb{Q})$ stable if
it is weak $(\lambda
,(\delta_{n,p_{n}})_{n,p_{n}},d)$ stable and if:%
\[
\Pr(\sup_{U_{n}\in\text{support}(\mathbb{Q)}}\frac{d(\psi_{U_{n}},\psi_{n}%
)}{||\mathbb{P}_{n,U_{n}}-\mathbb{P}_{n}||}\geq\lambda)\leq\delta_{n,p_{n}}.%
\]
\end{definition}

\noindent As before, we also define the following stronger notion:

\begin{definition}
[Cross-validation strong stability]Let $z\in\mathcal{Z}$. Let $\mathcal{D}%
_{n}=\mathcal{D}_{n-1}\cup\{z\}$ a learning set. Let $V_{n}^{tr}$ a
cross-validation vector with distribution $\mathbb{Q}$. A
learning algorithm
$\Psi$ is said to be strongly $(\lambda,(\delta_{n,p_{n}})_{n,p_{n}%
},d,\mathbb{Q})$ stable if it is strong $(\lambda,(\delta_{n,p_{n}})_{n,p_{n}%
},d)$ stable and if:%
\[
\Pr(\sup_{U_{n}\in\text{support}(\mathbb{Q)}}\frac{d(\psi_{U_{n}},\psi_{n}%
)}{||\mathbb{P}_{n,U_{n}}-\mathbb{P}_{n}||}\geq\lambda)\leq\delta_{n,p_{n}}.%
\]
\end{definition}

\begin{remark}
\label{Rem1}If the cardinal of the support of $\mathbb{Q}$ is denoted
$\kappa(n)$, then a learning algorithm which is weak $(\lambda,(\delta_{n,p_{n}%
})_{n,p_{n}},d,\mathbb{Q})$-stable is also strong $(\lambda,(\kappa
(n)\delta_{n,p_{n}})_{n},d,\mathbb{Q})$-stable.
\end{remark}

\noindent As seen in the following table, we retrieve with those notations the
different notions of stability introduced by \cite{DEWA79}, \cite{KEA95} and
also \cite{BE01}, \cite{KUNIY02}.%

\[%
\begin{tabular}
[c]{|l|l|l|l|}\hline
{\tiny stability distance} & ${\tiny d}_{\infty}$ & ${\tiny d}_{1}$ &
${\tiny d}_{e}$\\\hline
{\tiny Weak} &
\begin{tabular}
[c]{l}%
{\tiny weak }${\tiny (\lambda,\delta)}$ {\tiny hypothesis stability}\\
{\tiny \cite{KUNIY02}}%
\end{tabular}
&
\begin{tabular}
[c]{l}%
{\tiny weak }${\tiny (\lambda,\delta)}$ ${\tiny L}_{{\tiny 1}}$%
{\tiny stability}\\
{\tiny \cite{KUNIY02}}%
\end{tabular}
&
\begin{tabular}
[c]{l}%
{\tiny weak }${\tiny (\lambda,\delta)}$ {\tiny error stability}\\
{\tiny \cite{KUNIY02}}%
\end{tabular}
\\\hline
{\tiny Strong} &
\begin{tabular}
[c]{l}%
{\tiny strong }${\tiny (\lambda,\delta)}$ {\tiny hypothesis stability}\\
{\tiny \cite{KUNIY02}\cite{DEWA79}}%
\end{tabular}
&
\begin{tabular}
[c]{l}%
{\tiny strong }${\tiny (\lambda,\delta)}$ ${\tiny L}_{{\tiny 1}}%
${\tiny stability}\\
{\tiny \cite{KUNIY02}}%
\end{tabular}
&
\begin{tabular}
[c]{l}%
{\tiny strong }${\tiny (\lambda,\delta)}$ {\tiny error stability}\\
{\tiny \cite{KUNIY02}}%
\end{tabular}
\\\hline
{\tiny Sure Stability} &
\begin{tabular}
[c]{l}%
{\tiny uniform stability}\\
{\tiny \cite{BE01}}%
\end{tabular}
& {\tiny \cite{DEWA79}} &
\begin{tabular}
[c]{l}%
{\tiny error stability}\\
{\tiny \cite{KEA95}}%
\end{tabular}
\\\hline
\end{tabular}
\ \
\]

\noindent To motivate this approach, we also quote a list of class of
predictors satisfying the previous stability conditions.%

\[%
\begin{tabular}
[c]{|l|l|l|l|}\hline {\small stability distance} & ${\small
d}_{\infty}$ & ${\small d}_{1}$ & ${\small d}_{e}$\\\hline
{\small Weak} &  &  & {\small Lasso}%
\\\hline
{\small Strong} &  {\small Adaboost (}{\tiny \cite{KUNIY02})} &
\begin{tabular}
[c]{l}%
{\small -ERM (}{\tiny \cite{KUNIY02})}\\
-$k${\small -nearest rule}%
\end{tabular}
&
\begin{tabular}
[c]{l}%
{\small Bayesian algorithm}\\
{\small \cite{KEA95}}%
\end{tabular}
\\\hline
{\small Uniform} & {\small Regularization networks } &  &
\\\hline
\end{tabular}
\ \
\]

\bigskip

\noindent We recall the main notations and definitions:

\begin{table}[th]
\begin{center}%
\begin{tabular}
[c]{lll}\hline
Name & Notation & Definition\\\hline\hline
Risk or generalization error & $\widetilde{R}_{n}$ & $E_{P}[L(Y,\phi
(X,D_{n}))\mid D_{n}]$\\
Resubstitution error & $\widehat{R}_{n}$ & $\frac{1}{n}\sum_{i=1}^{n}%
L(Y_{i},\phi_{n}(X_{i},D_{n}))$\\
Cross-validation error & $\widehat{R}_{CV}$ & $E_{V_{n}^{tr}}P_{n,V_{n}^{ts}%
}\psi_{V_{n}^{tr}}$\\\hline
\end{tabular}
\end{center}
\caption{Main notations}%
\end{table}

\subsubsection{Main results}

\noindent Let $\mathcal{D}_{n}$ be a learning set of size $n$. Let $V_{n}%
^{tr}\sim$ $\mathbb{Q}$ be a training vector independent of $\mathcal{D}_{n}$
such that the cross-validation is symmetric and the number of elements in the
training set is constant and equal to $np_{n}$. Let $d$ be a distance among
$d_{e},d_{1},d_{\infty}$. At last, we suppose that the loss function $L$ is
bounded by $1$. We derive the following general results that stands for
general cross-validation procedures and stable algorithms.

\bigskip

\begin{theorem}
[Cross-validation Strong stability]\label{th:CV strong}Suppose that
$\mathcal{H}$ holds. Let $\Psi$ a machine learning which is strong
$(\lambda,(\delta_{n},p_{n})_{n,p_{n}},\mathbb{Q})$ stable with respect to the
distance $d$. Then, for all $\varepsilon\geq0$, we have:%

\[
\Pr\mathbb{(}\widetilde{R}_{n}(\Phi_{n}^{B})-\hat{R}_{CV}^{Out}\geq
\varepsilon)\leq\exp(-2np_{n}\varepsilon^{2})
\]

\noindent Furthermore, if $d$ is the uniform distance $d_{\infty}$, then we
have for all $\alpha>0$:%

\[
\Pr\mathbb{(}\widetilde{R}_{n}(\Phi_{n}^{B})-\hat{R}_{CV}^{Out}\geq
\varepsilon)\leq\min(\exp(-2np_{n}\varepsilon^{2}),2(\exp(-\frac{\varepsilon
^{2}}{8n(8\lambda np_{n}+\alpha)^{2}})+\frac{n}{\alpha}\delta_{n,p_{n}}))
\]

\noindent Thus, if we choose $\alpha=8\lambda np_{n}$,
\[
\Pr\mathbb{(}\widetilde{R}_{n}(\Phi_{n}^{B})-\hat{R}_{CV}^{Out}\geq
\varepsilon)\leq\min(\exp(-2np_{n}\varepsilon^{2}),2(\exp(-\frac{\varepsilon
^{2}}{8(16\lambda)^{2}np_{n}^{2}})+\frac{n}{8\lambda p_{n}}\delta_{n,p_{n}}))
\]
\end{theorem}

\noindent{\bfseries      Proof.}

\noindent On the one hand, we have as before by conditional
Hoeffding's inequality (for a proof, see e.g. chapter 1):
\[
\Pr(\widetilde{R}_{n}(\Phi_{n}^{B})-\hat{R}_{CV}^{Out}\geq\varepsilon)\leq
\Pr(\mathbb{E}_{V_{n}^{tr}}(\mathbb{P}\psi_{V_{n}^{tr}}\mathbb{-P}%
_{n,V_{n}^{ts}}\psi_{V_{n}^{tr}})\geq\varepsilon)\leq\exp(-2np_{n}%
\varepsilon^{2})
\]

\noindent On the other hand, notice that $\mathbb{P}^{\otimes n}%
\mathbb{E}_{V_{n}^{tr}}(\mathbb{P}\psi_{V_{n}^{tr}}\mathbb{-P}_{n,V_{n}^{ts}%
}\psi_{V_{n}^{tr}})=0$

\bigskip

\noindent Denote $f(Z_{1},Z_{2},\ldots,Z_{n}):=\mathbb{E}_{V_{n}^{tr}%
}(\mathbb{P}\psi_{V_{n}^{tr}}\mathbb{-P}_{n,V_{n}^{ts}}\psi_{V_{n}^{tr}})$.
Let $z$ $\in$ $\mathcal{Z}$. Now denote:
\[
B:=\{\sup_{U_{n}\in\text{support}(\mathbb{Q)}}\frac{d(\psi_{U_{n}},\psi
_{n+1})}{||\mathbb{P}_{n,U_{n}}-\mathbb{P}_{n+1}||}\geq\lambda\}
\]
\noindent with $\psi_{n+1}$ trained on $\mathcal{D}_{n+1}=\{Z_{1}%
,\ldots,Z_{i-1},Z_{i},Z_{i+1},\ldots,Z_{n},z\}$. Under our
assumptions, we have $\Pr(B)\mathbb{\leq\delta}_{n+1,p_{n+1}}$.

\bigskip

\noindent We want to show that with high probability there exist constants
$c_{i}$ such that for all $i\in\{1,\ldots,n\}$, for all $z\in\mathcal{Z},$
$$\Delta_i:=|f(Z_{1},\ldots,Z_{i},\ldots,Z_{n})-f(Z_{1},\ldots,Z_{i-1},z,Z_{i+1}%
,\ldots,Z_{n})|\leq c_{i}$$.

\noindent Notice that:%

\begin{align*}
|\Delta_i|  &
=|\mathbb{E}_{V_{n}^{tr}}(\mathbb{P}\psi_{V_{n}^{tr}}\mathbb{-P}_{n,V_{n}%
^{ts}}\psi_{V_{n}^{tr}})-(\mathbb{E}_{V_{n}^{tr}}\mathbb{P}\psi_{V_{n}^{tr}%
}^{^{\prime}}-\mathbb{P}_{n,V_{n}^{ts}}^{^{\prime}}\mathbb{P}\psi_{V_{n}^{tr}%
}^{^{\prime}})|\\
&  \leq|\mathbb{E}_{V_{n}^{tr}}\mathbb{P(}\psi_{V_{n}^{tr}}-\psi_{V_{n}^{tr}%
}^{^{\prime}})|+|\mathbb{E}_{V_{n}^{tr}}(\mathbb{P}_{n,V_{n}^{ts}}\psi
_{V_{n}^{tr}}\mathbb{-P}_{n,V_{n}^{ts}}^{^{\prime}}\mathbb{P}\psi_{V_{n}^{tr}%
}^{^{\prime}})|
\end{align*}

\noindent with $\mathbb{P}_{n,V_{n}^{tr}}^{^{\prime}}$ the weighted empirical
measure on the sample
$$\mathcal{E}_{n}=\{Z_{1},\ldots,Z_{i-1},z,Z_{i+1}%
,\ldots,Z_{n}\}$$

\noindent and $\psi_{V_{n}^{tr}}^{^{\prime}}$ \noindent the
predictor trained on $\mathcal{E}_{V_{n}^{tr}}$.

\bigskip

\noindent So, first, let us bound the first term, $|\mathbb{E}_{V_{n}^{tr}%
}\mathbb{P(}\psi_{V_{n}^{tr}}-\psi_{V_{n}^{tr}}^{^{\prime}})|\leq
\mathbb{E}_{V_{n}^{tr}}|\mathbb{P(}\psi_{V_{n}^{tr}}-\psi_{n+1})|+\mathbb{E}%
_{V_{n}^{tr}}|\mathbb{P(}\psi_{n+1}-\psi_{V_{n}^{tr}}^{^{\prime}})|$. Thus, on
$B^{\subset}$, we have $|\mathbb{E}_{V_{n}^{tr}}\mathbb{P(}\psi_{V_{n}^{tr}%
}-\psi_{V_{n}^{tr}}^{^{\prime}})|\leq\frac{4\lambda}{n+1}$.

\bigskip

\noindent To upper bound the second term, notice that:

\begin{align*}
|\mathbb{E}_{V_{n}^{tr}}\mathbb{P}_{n,V_{n}^{ts}}\psi_{V_{n}^{tr}}%
-\mathbb{E}_{V_{n}^{tr}}\mathbb{P}_{n,V_{n}^{ts}}^{^{\prime}}\psi_{V_{n}^{tr}%
}^{^{\prime}}|& =|\mathbb{E}_{V_{n}^{tr}}(\mathbb{P}_{n,V_{n}^{ts}}(\psi
_{V_{n}^{tr}}-\psi_{V_{n}^{tr}}^{^{\prime}})|V_{n,i}^{tr}=1)\times(1-p_{n})\\
&+\mathbb{E}_{V_{n}^{tr}}((\mathbb{P}_{n,V_{n}^{ts}}-\mathbb{P}_{n,V_{n}^{ts}%
}^{^{\prime}})\psi_{V_{n}^{tr}}|V_{n,i}^{ts}=1)\times p_{n}|
\end{align*}

\noindent We always have for any $\psi$, $|(\mathbb{P}_{n,V_{n}^{ts}%
}-\mathbb{P}_{n,V_{n}^{ts}}^{^{\prime}})\psi|\leq1/np_{n}$ thus $|\mathbb{E}%
_{V_{n}^{tr}}((\mathbb{P}_{n,V_{n}^{ts}}-\mathbb{P}_{n,V_{n}^{ts}}^{^{\prime}%
})\psi_{V_{n}^{tr}},V_{n}^{ts}=1)\times p_{n}|\leq1/n$

\bigskip

\noindent We still have to bound $|\mathbb{E}_{V_{n}^{tr}}(\mathbb{P}%
_{n,V_{n}^{ts}}(\psi_{V_{n}^{tr}}-\psi_{V_{n}^{tr}}^{^{\prime}})|V_{n,i}%
^{tr}=1)|$ which is always smaller than
$\mathbb{E}_{V_{n}^{tr}}(d_{\infty
}(\psi_{V_{n}^{tr}},\psi_{V_{n}^{tr}}^{^{\prime}})|V_{n,i}^{tr}=1)$
in the special case of the most stable kind of stability namely
the uniform stability.

\bigskip

\noindent On $B^{\subset}$, we get $d_{\infty}(\psi_{V_{n}^{tr}},\psi
_{V_{n}^{tr}}^{^{\prime}})\leq d_{\infty}(\psi_{V_{n}^{tr}},\psi
_{n+1})+d_{\infty}(\psi_{n+1},\psi_{V_{n}^{tr}}^{^{\prime}})\leq4\lambda
p_{n}$.

\noindent Thus, on $B^{\subset}$, we derive
$$\mathbb{E}_{V_{n}^{tr}}%
(d_{\infty}(\psi_{V_{n}^{tr}},\psi_{V_{n}^{tr}}^{^{\prime}})|V_{n,i}%
^{tr}=1)\leq4\lambda p_{n}.$$

\bigskip

\noindent Putting all together, with probability at least $1-\delta_{n,p_{n}}%
$, we get
\[
\sup_{1\leq i\leq n,z\in\mathcal{Z}}|f(Z_{1},\ldots,Z_{i},\ldots
,Z_{n})-f(Z_{1},\ldots,z,\ldots,Z_{n})|\leq\frac{4\lambda}{n+1}+4\lambda
p_{n}(1-p_{n})\leq8\lambda p_{n}.
\]

\noindent Applying theorem \ref{ch2:Kutin strong}, we obtain that
for all $\varepsilon \geq0$:
\begin{align*}
\Pr(\mathbb{E}_{V_{n}^{tr}}(\mathbb{P}\psi_{V_{n}^{tr}}\mathbb{-P}%
_{n,V_{n}^{ts}}\psi_{V_{n}^{tr}})\geq\varepsilon)  &  \leq2(\exp
(-\frac{\varepsilon^{2}}{8n(8\lambda p_{n}+\alpha)^{2}})+\frac{n}{\alpha
}\delta_{n,p_{n}}^{^{\prime}})\\
&  \leq2(\exp(-\frac{\varepsilon^{2}}{8(16\lambda)^{2}np_{n}^{2}}%
)+\frac{n}{8\lambda p_{n}}\delta_{n,p_{n}}^{^{\prime}})\text{ by taking
}\alpha=8\lambda p_{n}%
\end{align*}

$\Box$

\begin{theorem}
[Cross-validation Weak stability]Suppose that $\mathcal{H}$ holds. Let $\Psi$
be a machine learning which is weak\textbf{\ }$(\lambda,(\delta_{n,p_{n}%
})_{n,p_{n}},\mathbb{Q})$ stable with respect to the distance $d$. Then, for
all $\varepsilon\geq0$, we have%

\[
\Pr\mathbb{(}\widetilde{R}_{n}(\Phi_{n}^{B})-\hat{R}_{CV}^{Out}\geq
\varepsilon)\leq\exp(-2np_{n}\varepsilon^{2}).
\]

\noindent Furthermore, if the distance is the uniform distance $d_{\infty}$,
we have for all $\varepsilon\geq0$:

$\Pr\mathbb{(}\widetilde{R}_{n}(\Phi_{n}^{B})-\hat{R}_{CV}^{Out}%
\geq\varepsilon)\leq\min(\exp(-2np_{n}\varepsilon^{2}),2(\exp
(-\frac{n\varepsilon^{2}}{10(9\lambda np_{n})^{2}}+\frac{n\delta_{n(1-p_{n}%
)}^{1/2}}{9\lambda p_{n}}\exp(\frac{\varepsilon n}{4(9\lambda np_{n})^{2}%
}))+n\delta_{n,p_{n}}^{1/2}).$
\end{theorem}

\bigskip

{\bfseries   Proof.}

\noindent Denote $f(Z_{1},Z_{2},\ldots,Z_{n}):=\hat{R}_{CV}^{Out}%
-\widetilde{R}_{n}$ and $B:=\{\sup_{U_{n}\in\text{support}(\mathbb{Q)}%
}\frac{d(\psi_{U_{n}},\psi_{n+1})}{||\mathbb{P}_{n,U_{n}}-\mathbb{P}_{n+1}%
||}\geq\lambda\}$ with $\psi_{n+1}$ trained on $\mathcal{D}_{n+1}%
=\{Z_{1},\ldots,Z_{i-1},Z_{i},Z_{i+1},\ldots,Z_{n},Z_{i}^{^{\prime}}\}$.

\noindent We want to show that for all $i$, there exists constant $c_{i}$ such
$|\Delta_i|:=|f(Z_{1},\ldots,Z_{i},\ldots,Z_{n})-f(Z_{1},\ldots,Z_{i}^{^{\prime}}%
,\ldots,Z_{n})|\leq c_{i}$ with high probability where $Z_{1},\ldots
,Z_{i},\ldots,Z_{n},Z_{i}^{^{\prime}}$ are i.i.d. variables.%

\begin{align*}
|\Delta_i|  &  =|\mathbb{E}_{V_{n}^{tr}}(\mathbb{P}\psi_{V_{n}^{tr}%
}\mathbb{-P}_{n,V_{n}^{ts}}\psi_{V_{n}^{tr}})-(\mathbb{E}_{V_{n}^{tr}%
}\mathbb{P}\psi_{V_{n}^{tr}}^{^{\prime}}-\mathbb{P}_{n,V_{n}^{ts}}^{^{\prime}%
}\mathbb{P}\psi_{V_{n}^{tr}}^{^{\prime}})|\\
&  \leq\mathbb{E}_{V_{n}^{tr}}|\mathbb{P(}\psi_{V_{n}^{tr}}-\psi_{V_{n}^{tr}%
}^{^{\prime}})|+\mathbb{E}_{V_{n}^{tr}}|(\mathbb{P}_{n,V_{n}^{ts}}\psi
_{V_{n}^{tr}}\mathbb{-P}_{n,V_{n}^{ts}}^{^{\prime}}\mathbb{P}\psi_{V_{n}^{tr}%
}^{^{\prime}})|.
\end{align*}

\noindent with $\mathbb{P}_{n}^{^{\prime}},\mathbb{P}_{n,V_{n}^{ts}}%
^{^{\prime}}$ the weighted empirical measures of the sample $\mathcal{D}%
_{n}^{^{\prime}}=\{Z_{1},\ldots,Z_{i}^{^{\prime}},\ldots,Z_{n}\}$ and
$\psi_{n}^{^{\prime}}$ the predictor built on $\mathcal{D}_{n}^{^{\prime}}$.

\bigskip

\noindent So, first, let us bound the first term, $|\mathbb{E}_{V_{n}^{tr}%
}\mathbb{P(}\psi_{V_{n}^{tr}}-\psi_{V_{n}^{tr}}^{^{\prime}})|\leq
\mathbb{E}_{V_{n}^{tr}}|\mathbb{P(}\psi_{V_{n}^{tr}}-\psi_{n+1})|+\mathbb{E}%
_{V_{n}^{tr}}|\mathbb{P(}\psi_{n+1}-\psi_{V_{n}^{tr}}^{^{\prime}})|$ Thus, on
$B^{\subset}$, we have $|\mathbb{E}_{V_{n}^{tr}}\mathbb{P(}\psi_{V_{n}^{tr}%
}-\psi_{V_{n}^{tr}}^{^{\prime}})|\leq\frac{4\lambda}{n+1}.$

\noindent To upper bound the second term, notice that:%

\begin{align*}
{\small |}\mathbb{E}_{V_{n}^{tr}}{\small P}_{n,V_{n}^{ts}}{\small \psi}%
_{V_{n}^{tr}}{\small -}\mathbb{E}_{V_{n}^{tr}}{\small P}_{n,V_{n}^{ts}%
}^{^{\prime}}{\small \psi}_{V_{n}^{tr}}^{^{\prime}}{\small |}  &
{\small =|}\mathbb{E}_{V_{n}^{tr}}{\small (P}_{n,V_{n}^{ts}}{\small (\psi
}_{V_{n}^{tr}}{\small -\psi}_{V_{n}^{tr}}^{^{\prime}}{\small ),V}_{n}%
^{tr}{\small =1)\times(1-p}_{n}{\small )}\\
&  {\small +}\mathbb{E}_{V_{n}^{tr}}{\small ((P}_{n,V_{n}^{ts}}{\small -P}%
_{n,V_{n}^{ts}}^{^{\prime}}{\small )\psi}_{V_{n}^{tr}}{\small ,V}_{n}%
^{ts}{\small =}{\small 1)\times p}_{n}{\small |}.%
\end{align*}

\noindent We always have for all $\psi$, $|(\mathbb{P}_{n,V_{n}^{ts}%
}-\mathbb{P}_{n,V_{n}^{ts}}^{^{\prime}})\psi|\leq1/np_{n}$ thus we get
$$|\mathbb{E}_{V_{n}^{tr}}((\mathbb{P}_{n,V_{n}^{ts}}-\mathbb{P}_{n,V_{n}^{ts}%
}^{^{\prime}})\psi_{V_{n}^{tr}},V_{n}^{ts}=1)\times p_{n}|\leq1/n .$$

\bigskip

\noindent We still have to bound $|\mathbb{E}_{V_{n}^{tr}}(\mathbb{P}%
_{n,V_{n}^{ts}}(\psi_{V_{n}^{tr}}-\psi_{V_{n}^{tr}}^{^{\prime}}),V_{n}%
^{tr}=1)|\leq\mathbb{E}_{V_{n}^{tr}}(d_{\infty}(\psi_{V_{n}^{tr}},\psi
_{V_{n}^{tr}}^{^{\prime}}),V_{n}^{tr}=1)$ in the special of the uniform stability.

\bigskip

\noindent On $B^{\subset}$, we derive $d_{\infty}(\psi_{V_{n}^{tr}}%
,\psi_{V_{n}^{tr}}^{^{\prime}})\leq d_{\infty}(\psi_{V_{n}^{tr}},\psi
_{n+1})+d_{\infty}(\psi_{n+1},\psi_{V_{n}^{tr}}^{^{\prime}})\leq4\lambda
p_{n}$, thus on $B^{\subset}$
$$\mathbb{E}_{V_{n}^{tr}}(d_{\infty}(\psi_{V_{n}^{tr}},\psi
_{V_{n}^{tr}}^{^{\prime}}),V_{n}^{tr}=1)\leq4\lambda p_{n}.$$

\bigskip

\noindent Putting all together, with probability at least $1-\delta_{n,p_{n}}%
$,
$$|f(Z_{1},\ldots,Z_{i},\ldots,Z_{n})-f(Z_{1},\ldots,Z_{i^{\prime}}%
,\ldots,Z_{n})|\leq8\lambda p_{n}.$$

$\Box$

\bigskip

\noindent Following the previous results, we can obtain results for the
expectation of the difference $\widetilde{R}_{n}(\Phi_{n}^{B})-\hat{R}%
_{CV}^{Out}$.

\begin{theorem}
In the case of classification, we can bound the excess risk by%

\[
\mathbb{E}_{\mathcal{D}_{n}}(\widetilde{R}_{n}(\Phi_{n}^{B})-\hat{R}%
_{CV}^{Out})\leq\sqrt{1/np_{n}}%
\]

\noindent Furthermore, if $d$ is the uniform distance $d_{\infty}$, then we
have for all $\alpha>0$:%

\[
\mathbb{E}_{\mathcal{D}_{n}}(\widetilde{R}_{n}(\Phi_{n}^{B})-\hat{R}%
_{CV}^{Out})\leq\min(\sqrt{1/np_{n}},\sqrt{16^{3}n}\lambda p_{n}%
+\frac{n}{4\lambda p_{n}}\delta_{n,p_{n}})
\]
\end{theorem}

\noindent Similar results can be derived in the context of the weak stability.

\noindent{\bfseries      Proof}

\noindent It is sufficient to apply the previous probability upper bounds
together with the lemma \ref{lem esperance}.

\bigskip

$\Box$

\section{Results for the cross-validated subagged classification}

\noindent In the case of subagging of classifiers (i.e. the majority vote), we
can obtain the following results:

\begin{theorem}
For any subbaged classifier, we can bound the excess risk.%

\[
\Pr(\widetilde{R}_{n}(\Phi_{n}^{B})-\frac{1}{2}\hat{R}_{CV}^{Out}%
\geq\varepsilon)\leq\exp(-8np_{n}\varepsilon^{2}/9)
\]

\noindent and also
\[
\Pr(\widetilde{R}_{n}(\Phi_{n}^{B})-l\hat{R}_{CV}^{Maj}\geq\varepsilon)\leq
l\exp(-2np_{n}\varepsilon^{2}/9)
\]

\noindent where $N$ denotes the total number of training vectors
in the cross-validation and $l$ denotes $[(N-1)/2]+1$ that is the
strict majority of the subbaged classifiers and
$\hat{R}_{CV}^{Maj}$ the cross-validated estimate of this
majority.

\noindent Furthermore, in the particular case of binary classification we also have%

\[
\Pr(\widetilde{R}_{n}(\Phi_{n}^{B})-(\hat{R}_{CV}^{Out}/2-1/2))\leq
-\varepsilon)\leq\exp(-2np_{n}\varepsilon^{2}/9)
\]

and%
\[
\Pr(\widetilde{R}_{n}(\Phi_{n}^{B})-(l\hat{R}_{CV}^{Maj}-l+1)\leq
-\varepsilon)\leq l\exp(-2np_{n}\varepsilon^{2})
\]
\end{theorem}

\noindent{\bfseries   Proof.}

\noindent We consider a ghost sample i.i.d. of size $m$: $(X_{1}^{^{\prime}%
},Y_{1}^{^{\prime}}),...,(X_{m}^{^{\prime}},Y_{m}^{^{\prime}})$.
Denote
$\eta_{i}:=L(Y_{i}^{^{\prime}},\phi_{n}^{B}(X_{i}^{^{\prime}}))$.

\noindent Then $e_{m}^{B}:=\frac{1}{m}\sum_{i=1}^{m}\eta_{i}$ corresponds to
the average number of mistakes of $\Phi_{n}^{B}$ on the ghost sample. In the
same way, $e_{m}^{_{V_{n}^{tr}}}:=\frac{1}{m}\sum_{i=1}^{m}L(Y_{i}^{^{\prime}%
},\phi_{V_{n}^{tr}}(X_{i}^{^{\prime}}))$ (respectively $e_{m}%
^{a}:=\mathbb{E}_{V_{n}^{tr}}[\frac{1}{m}\sum_{i=1}^{m}L(Y_{i}^{^{\prime}%
},\phi_{V_{n}^{tr}}(X_{i}^{^{\prime}}))]$) is the average number
of the mistakes of $\phi_{V_{n}^{tr}}$ (respectively the weighted
average number of mistakes of the family of predictors
$\phi_{V_{n}^{tr}}$).

\bigskip

\bigskip

Denote by

\begin{enumerate}
\item $L_{1}:=\widetilde{R}_{n}(\Phi_{n}^{B})-\frac{1}{2}\hat{R}_{CV}^{Out}$

\item $L_{2}:=\widetilde{R}_{n}(\Phi_{n}^{B})-e_{m}^{B}$

\item $L_{3}:=e_{m}^{B}-e_{m}^{a}/2$

\item $L_{4}:=\frac{1}{2}[e_{m}^{a}-E_{X,Y}\mathbb{E}_{V_{n}^{tr}}%
L(Y,\phi_{V_{n}^{tr}}(X))]$

\item $L_{5}:=\frac{1}{2}[E_{X,Y}\mathbb{E}_{V_{n}^{tr}}L(Y,\phi
_{V_{n}^{tr}}(X))-\hat{R}_{CV}^{Out}]$
\end{enumerate}

\bigskip

\noindent We have
\[
\Pr(L_{1}\geq3\varepsilon)\leq\Pr(L_{2}\geq\varepsilon)+\Pr(L_{3}\geq
0)+\Pr(L_{4}\geq\varepsilon)+\Pr(L_{5}\geq\varepsilon)
\]

\bigskip

\noindent By Hoeffding's inequality, we have:%
\[
\Pr(L_{2}\geq\varepsilon)\leq\exp(-2m\varepsilon^{2}).
\]

\noindent and also $\Pr(L_{4}\geq\varepsilon)\leq\exp(-2m(2\varepsilon)^{2})$

\bigskip

\noindent By conditionnal Hoeffding's inequality (for a proof, see e.g.
\cite{COR09A}), we deduce
\[
\Pr(L_{5}\geq\varepsilon)\leq\exp(-2np_{n}(2\varepsilon)^{2})
\]

\bigskip

\noindent By conditionnal Hoeffding's inequality, we also have%

\[
\Pr(e_{m}^{a}-E_{X,Y}\mathbb{E}_{V_{n}^{tr}}L(Y,\phi_{V_{n}^{tr}}%
(X))\geq\varepsilon)\leq\exp(-2m\varepsilon^{2}).
\]

\noindent since for fixed $v_{n}^{tr}$ $\Pr(\frac{1}{m}\sum_{i=1}^{m}%
L(Y_{i}^{^{\prime}},\phi_{v_{n}^{tr}}(X_{i}^{^{\prime}}))-E_{X,Y}%
L(Y,\phi_{v_{n}^{tr}}(X))\geq\varepsilon)\leq\exp(-2m\varepsilon^{2})$

\bigskip

\noindent We suppose here that $\Pr(V_{n}^{tr}=v_{n})$ are
rational numbers whose smallest multiplicator is denoted by $N$.
Thus $e_{m}^{a}$ can be seen as a simple average number of
mistakes of a family of predictors $(\phi _{j})_{1\leq j\leq N}$
on the ghost sample.

\bigskip

\noindent First notice, that if $e_{m}^{a}$ is small then $e_{m}^{B}$ must be
small either. Indeed,we have
\[
e_{m}^{a}=\frac{1}{N}\sum_{j=1}^{N}\frac{1}{m}\sum_{i=1}^{m}L(Y_{i}^{^{\prime
}},\phi_{j}(X_{i}^{^{\prime}}))=\frac{1}{N}\frac{1}{m}\sum_{1\leq
j\leq
N,1\leq i\leq m}^{N}\epsilon_{i,j}%
\]

\noindent with $\epsilon_{i,j}:=L(Y_{i}^{^{\prime}},\phi_{j}(X_{i}%
^{^{\prime}}))\in\{0,1\}.$\ We thus deduce that the total number
of mistakes on the ghost sample of the family of predictors
$(\phi_{j})_{1\leq j\leq N}$ is equal to $Nme_{m}^{a}$. Notice
that if the number of mistakes of the family $(\phi_{j})_{1\leq
j\leq N}$ on the $i$-th observation is less that
$\lfloor(N-1)/2\rfloor$ (i.e.
$\sum_{j=1}^{N}\epsilon_{i,j}\leq\lfloor (N-1)/2\rfloor$) then it
means that a strict majority of predictors have classified
correctly $Y_{i}^{^{\prime}}$, which in turns tells us that a
strict majority of predictors have the same output $Y_{i}^{^{\prime}}%
=\phi_{j}(X_{i}^{^{\prime}})$. We thus have $\phi_{n}^{B}(X_{i}%
^{^{\prime}})=Y_{i}^{^{\prime}}$ which implies $\eta_{j}=L(Y_{i}^{^{\prime}%
},\phi_{n}^{B}(X_{i}^{^{\prime}}))=0$.

\bigskip

\noindent Denoting by $\kappa=me_{m}^{B}$ the number of mistakes of the
subbaged classifier on the ghost sample, we necessarly have
\[
\sum_{i=1}^{m}\sum_{j=1}^{N}\epsilon_{i,j}\geq\kappa(\lfloor(N-1)/2\rfloor
+1)=\kappa(\lfloor(N+1)/2\rfloor).
\]

\noindent It follows that%
\[
e_{m}^{B}\leq\frac{N}{\lfloor(N+1)/2\rfloor}e_{m}^{a}<e_{m}^{a}/2.
\]

\noindent Thus $\Pr(L_{3}\geq0)=0$

\bigskip

\noindent We conclude $\Pr(\widetilde{R}_{n}(\Phi_{n}^{B})-\frac{1}{2}\hat
{R}_{CV}^{Out}\geq 3 \varepsilon)\leq\exp(-2np_{n}(2\varepsilon)^{2}%
)+\exp(-2m(2\varepsilon)^{2})+\exp(-2m\varepsilon^{2}).$

\noindent If we let $m\rightarrow\infty$,
\[
\Pr(\widetilde{R}_{n}(\Phi_{n}^{B})-\frac{1}{2}\hat{R}_{CV}^{Out}%
\geq\varepsilon)\leq\exp(-8np_{n}\varepsilon^{2}/9)
\]

\bigskip

\noindent Notice that in the particular case of the binary classification, we
have by symmetry, $1-e_{m}^{B}\leq\frac{N}{\lfloor(N+1)/2\rfloor}(1-e_{m}%
^{a})$, which gives
\[
\frac{N}{\lfloor N/2+1\rfloor}e_{m}^{a}-(1-\frac{N}{\lfloor N/2+1\rfloor})\leq
e_{m}^{B}%
\]

\noindent and eventually $e_{m}^{B}\geq\frac{N}{\lfloor N/2+1\rfloor}e_{m}%
^{a}-1/2\geq e_{m}^{a}-1/2$

\bigskip

\noindent Thus, for binary classification, we can even obtain an probability
upper bound for $\Pr(|\widetilde{R}_{n}(\Phi_{n}^{B})-\frac{1}{2}\hat{R}%
_{CV}^{Out}|\geq\varepsilon)$ not only for $\Pr(\widetilde{R}_{n}(\Phi_{n}%
^{B})-\frac{1}{2}\hat{R}_{CV}^{Out}\geq\varepsilon)$. Indeed, denote by

\begin{enumerate}
\item $L_{1}^{^{\prime}}:=\widetilde{R}_{n}(\Phi_{n}^{B})-\frac{N}{\lfloor
N/2+1\rfloor}(\hat{R}_{CV}^{Out}-1/2)$

\item $L_{2}^{^{\prime}}:=\widetilde{R}_{n}(\Phi_{n}^{B})-e_{m}^{B}$

\item $L_{3}^{^{\prime}}:=e_{m}^{B}-(\frac{N}{\lfloor N/2+1\rfloor}e_{m}^{a}-1/2)$

\item $L_{4}^{^{\prime}}:=(\frac{N}{\lfloor N/2+1\rfloor}e_{m}^{a}%
-1/2)-(\frac{N}{\lfloor N/2+1\rfloor}E_{X,Y}\mathbb{E}_{V_{n}^{tr}}%
L(Y,\phi_{V_{n}^{tr}}(X))-1/2)$

\item $L_{5}^{^{\prime}}:=(\frac{N}{\lfloor N/2+1\rfloor}E_{X,Y}%
\mathbb{E}_{V_{n}^{tr}}L(Y,\phi_{V_{n}^{tr}}(X))-1/2)-(\frac{N}{\lfloor
N/2+1\rfloor}\hat{R}_{CV}^{Out}-1/2)$
\end{enumerate}

\bigskip

\noindent We get%

\begin{align*}
\Pr(L_{1}^{^{\prime}}  & \leq-3\varepsilon)\leq\Pr(L_{2}^{^{\prime}}%
\leq-\varepsilon)+\Pr(L_{3}^{^{\prime}}<0)+\Pr(L_{4}^{^{\prime}}%
\leq-\varepsilon)+\Pr(L_{5}^{^{\prime}}\leq-\varepsilon)\\
& \leq\exp(-2m\varepsilon^{2})+0+\exp(-2m(\frac{N}{\lfloor N/2+1\rfloor
}\varepsilon)^{2})+\exp(-2np_{n}(\frac{N}{\lfloor N/2+1\rfloor}\varepsilon
)^{2})
\end{align*}

\noindent Taking $m\rightarrow\infty$, and noticing that $N/\lfloor
N/2+1\rfloor>1$%

\begin{align*}
\Pr(\widetilde{R}_{n}(\Phi_{n}^{B})-(\hat{R}_{CV}^{Out}/2-1/2))\leq
-\varepsilon) & \leq\Pr(\widetilde{R}_{n}(\Phi_{n}^{B})-(\hat{R}_{CV}%
^{Out}/2-1/2)\leq-\varepsilon) \\
& \leq\Pr(L_{1}^{^{\prime}}\leq-\varepsilon
)\leq\exp(-2np_{n}\varepsilon^{2}/9)
\end{align*}

\bigskip

\noindent For binary classification, we can eventually obtain that
\[
\Pr(|\widetilde{R}_{n}(\Phi_{n}^{B})-\frac{1}{2}(\hat{R}_{CV}^{Out}%
-1/2)|\geq\varepsilon)\leq\exp(-8np_{n}\varepsilon^{2}/9)+\exp(-2np_{n}%
\varepsilon^{2}/9)\leq2\exp(-2np_{n}\varepsilon^{2}/9)
\]

\bigskip

\noindent Denote by $\epsilon_{j}:=\frac{1}{m}\sum_{i=1}^{m}\epsilon_{i,j}$
the average number of mistakes by predictors $j$ on the ghost sample. We can
order them by increasing order: $\epsilon_{(1)},...,\epsilon_{(N)}$. Let
$l:=\lfloor N/2+1\rfloor$ be the strict majority. An interesting case is when
we know that a strict majority of classifiers are very good. Denote by
\[
e_{m}^{G}:=\frac{1}{l}\sum_{j=1}^{l}\epsilon_{(j)}%
\]
\noindent their global average error of the first $l$ best classifiers on the
ghost sample.

\bigskip

\bigskip

In the same way, denote by $\mu_{j}:=E_{X,Y}L(Y,\phi_{j}(X))$ the
risk of the $j$-th classifier. We introduce now a
cross-validation estimate of the average risk
$\frac{1}{l}\sum_{j=1}^{l}\mu_{(j)}$ of the $l$ best classifiers:
$\hat{R}_{CV}^{Maj}$. For this, recall that each $\phi_{j}$
corresponds to some $\phi_{v_{n}^{tr}}$ thus we can define an out
sample error for the predictor
$j:\hat{r}_{j}:=\mathbb{P}_{n,v_{n}^{ts}}(L(Y,\phi_{j}(X))$. And
we define
$\hat{R}_{CV}^{Maj}:=\frac{1}{l}\sum_{j=1}^{l}\hat{r}_{(j)}$

\begin{enumerate}
\item $R_{1}:=\widetilde{R}_{n}(\Phi_{n}^{B})-l\hat{R}_{CV}^{Maj}$

\item $R_{2}:=\widetilde{R}_{n}(\Phi_{n}^{B})-e_{m}^{B}$

\item $R_{3}:=e_{m}^{B}-le_{m}^{G}$

\item $R_{4}:=l(e_{m}^{G}-\frac{1}{l}\sum_{j=1}^{l}\mu_{(j)})$

\item $R_{5}:=l(\frac{1}{l}\sum_{j=1}^{l}\mu_{(j)}-\hat{R}_{CV}^{Maj})$
\end{enumerate}

\noindent We have
\[
\Pr(R_{1}\geq3\varepsilon)\leq\Pr(R_{2}\geq\varepsilon)+\Pr(R_{3}>0)+\Pr
(R_{4}\geq\varepsilon)+\Pr(R_{5}\geq\varepsilon)
\]

\bigskip

\noindent By Hoeffding's inequality, we have:%
\[
\Pr(R_{2}\geq\varepsilon)\leq\exp(-2m\varepsilon^{2}).
\]

\bigskip

\noindent We also derive\bigskip

\bigskip$\Pr(R_{4}\geq\varepsilon)=\Pr(e_{m}^{G}-\frac{1}{l}\sum_{j=1}^{l}%
\mu_{(j)}\geq\varepsilon/l)=\Pr(\sum_{j=1}^{l}\epsilon_{(j)}-\sum_{j=1}^{l}%
\mu_{(j)}\geq\varepsilon)$

There exist permutations $\sigma$ and $\sigma^{^{\prime}}$ such that
$\epsilon_{(j)}=\epsilon_{\sigma(j)}$ and $\mu_{(j)}=\mu_{\sigma^{^{\prime}%
}(j)}$. Thus, we get
\begin{align*}
\Pr(R_{4}  & \geq\varepsilon)\leq\Pr(\sum_{j=1}^{l}\epsilon_{\sigma(j)}%
-\mu_{\sigma^{^{\prime}}(j)}\geq\varepsilon)\\
& \leq\Pr(\sum_{j=1}^{l}\epsilon_{\sigma^{^{\prime}}(j)}-\mu_{\sigma
^{^{\prime}}(j)}\geq\varepsilon)
\end{align*}

by definition of $\epsilon_{(j)}.$ It follows that
\begin{align*}
\Pr(R_{4}  & \geq\varepsilon)\leq\sum_{j=1}^{l}\Pr(\epsilon_{\sigma^{^{\prime
}}(j)}-\mu_{\sigma^{^{\prime}}(j)}\geq\varepsilon)\\
& \leq l\exp(-2m\varepsilon^{2}).
\end{align*}

\bigskip

\noindent In the same way, we deduce $\Pr(R_{5}\geq\varepsilon)\leq
l\exp(-2np_{n}\varepsilon^{2})$.

\noindent By conditional Hoeffding's inequality (for a proof, see
e.g.
\cite{COR09A}), we deduce $\Pr(L_{5}\geq\varepsilon)\leq\exp(-2np_{n}%
(2\varepsilon)^{2})$ and also for a fixed $v_{n}^{tr}$%

\[
\Pr(|e_{m}^{_{v_{n}^{tr}}}-E_{X,Y}L(Y,\phi_{v_{n}^{tr}}(X))|\geq
\varepsilon)\leq2\exp(-2m\varepsilon^{2}).
\]

\noindent By conditional Hoeffding's inequality (for a proof, see
e.g.
\cite{COR09A}), we also have%

\[
\Pr(|e_{m}^{a}-E_{X,Y}\mathbb{E}_{V_{n}^{tr}}L(Y,\phi_{V_{n}^{tr}}%
(X))|\geq\varepsilon)\leq2\exp(-2m\varepsilon^{2}).
\]

\noindent Notice that if all the $l$ best classifiers classify correctly the
$i$-th observation (i.e. $\epsilon_{i,(j)}=0$ for all $j\in\{1,...,M\}$), then
the subbaged classification classifies also correctly. Thus $\eta_{i}=0$. Let
$\kappa$ be the number of mistakes of the subbaged classifier on the ghost
sample and let $x$ the number of observations correctly classified by all the
$l$ classifiers. Then we obtain that the number of correctly classified
observations by the subagging is greater that $x$, i.e. $m-\kappa\geq x$. On
the other hand, there is at least one predictor that makes a mistake on each
of the remaining $m-x$ observations. Thus $m-x$ is less that the total number
of mistakes made by the $l$ best classifiers%

\[
(m-x)\leq mle_{m}^{G}.
\]

\noindent From which, it follows that
\[
e_{m}^{B}\leq le_{m}^{G}.
\]
\bigskip

\noindent Thus $\Pr(R_{3}>0)=0.$

\bigskip

\noindent Putting altogether, we have
\[
\Pr(\widetilde{R}_{n}(\Phi_{n}^{B})-l\hat{R}_{CV}^{Maj}\geq3\varepsilon
)\leq\exp(-2m\varepsilon^{2})+l\exp(-2m\varepsilon^{2})+l\exp(-2np_{n}%
\varepsilon^{2}).
\]

\noindent If we let $m\rightarrow\infty$, $\Pr(\widetilde{R}_{n}(\Phi_{n}%
^{B})-l\hat{R}_{CV}^{Maj}\geq\varepsilon)\leq l\exp(-2np_{n}\varepsilon^{2}/9).$

\bigskip

\noindent Once again, in the particular case of binary classification, we have
by symmetry $1-e_{m}^{B}\leq l(1-e_{m}^{G})$ which leads to%
\[
e_{m}^{B}\geq1-l(1-e_{m}^{G}).
\]

\noindent In the same way, we have a symmetrical result for binary
classification:%
\begin{align*}
\Pr(\widetilde{R}_{n}(\Phi_{n}^{B})-(l\hat{R}_{CV}^{Maj}-l+1)  &
\leq-3\varepsilon)\leq\exp(-2m\varepsilon^{2})+l\exp(-2m\varepsilon^{2}%
)+l\exp(-2np_{n}\varepsilon^{2})\\
& \leq l\exp(-2np_{n}\varepsilon^{2}).
\end{align*}

\noindent which gives $\Pr(|\widetilde{R}_{n}(\Phi_{n}^{B})-(l\hat{R}%
_{CV}^{Maj}-l+1)|\geq\varepsilon)\leq2l\exp(-2np_{n}\varepsilon^{2}/9).$

\bigskip

$\Box$

\bigskip

\noindent In the case of subagging of classifiers (i.e. the majority vote)
whose VC dimension is finite, we can obtain a stronger result:

\begin{theorem}
Suppose $\mathcal{H}$ holds and that the machine learning is based on
empirical risk minimization. We can bound the excess risk.%

\[
\Pr(\widetilde{R}_{n}(\Phi_{n}^{B})-\frac{1}{2}\hat{R}_{CV}^{Out}%
\geq\varepsilon)\leq\min(\exp(-8np_{n}\varepsilon^{2}/9),(2n(1-p_{n}%
)+1)^{4V_{\mathcal{C}}/(1-p_{n})}e^{-4n(1-p_{n})\varepsilon^{2}}).
\]

\noindent and also
\[
\Pr(\widetilde{R}_{n}(\Phi_{n}^{B})-l\hat{R}_{CV}^{Maj}\geq\varepsilon)\leq
l\exp(-2np_{n}\varepsilon^{2}/9)
\]

\noindent with the $l:=[(N-1)/2]+1$ the strict majority of the
subagged classifiers and $\hat{R}_{CV}^{Maj}$ the cross-validated
estimate of this majority.

\noindent Furthermore, in the particular case of binary classification we also have%

\[
\Pr(\widetilde{R}_{n}(\Phi_{n}^{B})-(\hat{R}_{CV}^{Out}/2-1/2))\leq
-\varepsilon)\leq\min(\exp(-2np_{n}\varepsilon^{2}/9),(2n(1-p_{n}%
)+1)^{4V_{\mathcal{C}}/(1-p_{n})}e^{-4n(1-p_{n})\varepsilon^{2}})
\]

and%
\[
\Pr(\widetilde{R}_{n}(\Phi_{n}^{B})-(l\hat{R}_{CV}^{Maj}-l+1)\leq
-\varepsilon)\leq l\exp(-2np_{n}\varepsilon^{2})
\]
\end{theorem}

\noindent{\bfseries   Proof.}
\bigskip

\noindent We use again the lemma (for a proof, see chapter 1): $\hat
{R}_{CV}^{Out}\geq\mathbb{P}_{n}L(Y,\phi_{n}(X))$ since
$$\phi_{n}=\arg
\min_{\phi\in\mathcal{C}}\frac{1}{n}\sum_{i=1}^{n}L(Y_{i},\phi(X_{i})).$$

\bigskip

Following the last proof, we can bound $L_{5}$ in another way.%

\begin{align*}
\Pr(L_{5}   \geq3\varepsilon) &\leq\Pr(\mathbb{E}_{V_{n}^{tr}}[E_{X,Y}%
L(Y,\phi_{V_{n}^{tr}}(X))-\mathbb{P}_{n}L(Y,\phi_{n}(X))]\geq
6\varepsilon)\\
& \leq\Pr(\mathbb{E}_{V_{n}^{tr}}[E_{X,Y}L(Y,\phi_{V_{n}^{tr}%
}(X))-\mathbb{P}_{n}L(Y,\phi_{n}(X))]\geq6\varepsilon)
\end{align*}

\bigskip

Then as in proof, we split according to $\mathbb{P}L(Y,\phi_{opt}(X))$ and we
obtain by lemma \ref{lem esperance}%

\[
\Pr(L_{5}\geq\varepsilon)\leq(2n(1-p_{n})+1)^{4V_{\mathcal{C}}/(1-p_{n}%
)}e^{-n(1-p_{n})(2\varepsilon)^{2}}%
\]

\bigskip

$\Box$

\section{Results for the subagged predictor selection}

\noindent The remaining important question is: in practice, how
should we choose $p_{n}$? We give a hint for this question.

\noindent First, suppose that the final user wants to have an accuracy equal
to a certain level $\eta$.
\bigskip

\noindent Then we need to provide him a rule to chose an
optimal $p_{n}^{\star}\,$ and to upper bound the probability of excess risk
$\Pr(\widetilde{R}_{n}(\phi_{n}^{B,p_{n}^{\star}})-\hat{R}_{CV}^{Out}%
(p_{n}^{\star})\geq\eta)$. Previous bounds tell us that for any fixed $p_{n}$,
$\Pr(\widetilde{R}_{n}(\phi_{n}^{B})-\hat{R}_{CV}^{Out}(p_{n}%
)\geq\varepsilon)\leq\min(B(n,p_{n},\varepsilon),V(n,p_{n},\varepsilon))$.
Notice that $\min(B(n,p_{n},\varepsilon),V(n,p_{n},\varepsilon))$ seen as a
function of $\varepsilon$ is a continuous non-increasing function.\ Thus, we
can define an inverse denoted by $f$.

\bigskip

\noindent The previous probability bound becomes
for any $p_{n}$: $\Pr(\widetilde{R}_{n}(\phi_{n}^{B})-\hat{R}_{CV}%
^{Out}(p_{n})\geq f(n,p_{n},\delta)\leq\delta$. For each $k$, define
$\delta_{n,k}$ by $f(n,k/n,\delta_{n,k})=\eta$, i.e. $\delta_{n,k}%
=\min(B(n,k/n,\eta),V(n,k/n,\eta))$. Denote $k_{n}^{\star}:=\arg\min
_{k\in\{1...n-1\}}\hat{R}_{CV}^{Out}(k/n)+f(n,k/n,\delta_{n,k})$ and denote by
$p_{n}^{\star}:=k_{n}^{\star}/n$. Thus, we obtain:

\begin{theorem}
[Subbaging selection]\noindent Suppose that $\mathcal{H}$ holds. Suppose also
that $\phi_{n}$ is based on empirical risk minimization. But instead of
minimizing $\widehat{R}_{n}(\phi)$, we suppose $\phi_{n}$ minimizes
$\frac{1}{n}\sum_{i=1}^{n}C(h(Y_{i},\phi(X_{i}))$. For simplicity, we suppose
the infimum is attained i.e. $\phi_{n}=\arg\min_{\phi\in\mathcal{C}}%
\frac{1}{n}\sum_{i=1}^{n}C(h(Y_{i},\phi(X_{i}))$. In this context, we have:

\begin{itemize}
\item if $\delta\geq\delta_{n}$
\[
f(n,p_{n},\delta)=\sqrt{\frac{\ln(1/\delta)}{2np_{n}}}%
\]

\item and if $\delta<\delta_{n}$,
\end{itemize}%

\[
f(n,p_{n},\delta)=3\sqrt{\frac{4V_{\mathcal{C}}\ln(2n(1-p_{n})+1)/(1-p_{n}%
)+\ln(1/\delta)}{n}}%
\]

\noindent with $\delta_{n}:=(2n(1-p_{n})+1)^{-\frac{4p_{n}V_{\mathcal{C}}%
}{(1-p_{n})(1/9-2p_{n})}}.$

\noindent Furthermore, we have for all $\varepsilon>0$:%

\[
\Pr(\widetilde{R}_{n}(\phi_{n}^{B,p_{n}^{\star}})-\hat{R}_{CV}^{Out}%
(p_{n}^{\star})\geq\varepsilon)=O_{n}((n+1)^{8V_{\mathcal{C}}}\exp
\left( -\frac{2n(\varepsilon-2\sqrt{2}V_{\mathcal{C}}^{1/2}\sqrt{\ln(n)/n})^{2}%
}{1-\exp(-2\varepsilon^{2})}\right)).
\]
\end{theorem}

\noindent{\bfseries      Proof}

\noindent We have:%
\begin{align*}
\Pr(\widetilde{R}_{n}(\phi_{n}^{B,p_{n}^{\star}})-\hat{R}_{CV}^{Out}%
(p_{n}^{\star})    \geq\eta) &=\Pr(\widetilde{R}_{n}(\phi_{n}^{B,p_{n}%
^{\star}})-\hat{R}_{CV}^{Out}(p_{n}^{\star})\geq f(n,p_{n}^{\star}%
,\delta_{n,k_{n}^{\star}}))\\
&  \leq\sum_{k\in\{1...n-1\}}\Pr(\widetilde{R}_{n}(\phi_{n}^{B,p_{k}}%
)\geq\hat{R}_{CV}^{Out}(p_{k})+f(n,k/n,\delta_{n,k})).
\end{align*}

\noindent It follows that:%
\begin{align*}
\Pr(\widetilde{R}_{n}(\phi_{n}^{B,p_{n}^{\star}})-\hat{R}_{CV}^{Out}%
(p_{n}^{\star})    \geq\eta) & \leq\sum_{k\in\{1...n-1\}}\Pr(\widetilde{R}%
_{n}(\phi_{n}^{B,p_{k}})-\hat{R}_{CV}^{Out}(p_{k})\geq\eta)\\
&  \leq\sum_{k\in\{1...n-1\}}\min(B(n,k/n,\eta),V(n,k/n,\eta)).
\end{align*}
\noindent Thus, using previous bounds we get:%

\begin{align*}
\Pr(\widetilde{R}_{n}(\phi_{n}^{B,p_{n}^{\star}})-\hat{R}_{CV}^{Out}%
(p_{n}^{\star})    \geq \eta) &\leq\min_{k_{0}\in\{1...n-1\}}(  \sum
_{k=1}^{k_{0}-1}(2n(1-k/n)+1)^{4V_{\mathcal{C}}/(1-k/n)}\exp(-2n\eta^{2}%
) \\
& \quad \quad +\sum_{k=k_{0}}^{n-1}\exp(-2k\eta^{2})) \\
&  \leq\min_{k_{0}\in\{1...n-1\}}(  k_{0}(2n+1)^{4V_{\mathcal{C}%
}/(1-k_{0}/n)}\exp(-2n\eta^{2}) \\
& \quad \quad +\exp(-2k_{0}\eta^{2})\frac{1-(
\exp(-2k\eta^{2}))  ^{n-k_{0}}}{1-\exp(-2\eta^{2})}) \\
&  \leq\min_{k_{0}\in\{1...n-1\}}(  (2n+1)^{4V_{\mathcal{C}}/(1-k_{0}%
/n)}\alpha^{n}+\frac{\alpha^{k_{0}}}{1-\alpha})  \text{ with }%
\alpha:=\exp(-2\eta^{2})
\end{align*}

\noindent We look for $k_{0}$ in $\{(1-z_{n})n,0<z_{n}<1$ and $z_{n}%
\rightarrow_{n\infty}0\}$%

\[
\Pr(\widetilde{R}_{n}(\phi_{n}^{B,p_{n}^{\star}})-\hat{R}_{CV}^{Out}%
(p_{n}^{\star})\geq\eta)\leq\min_{z_{n}}\left(  (2n+1)^{4V_{\mathcal{C}}%
/z_{n}}\alpha^{n}+\frac{\alpha^{(1-z_{n})n}}{1-\alpha}\right)
\]

\noindent We look for $z_{n}$ such that $(2n+1)^{4V_{\mathcal{C}}/z_{n}}%
\sim_{n\infty}\frac{\alpha^{-z_{n}n}}{1-\alpha}$

\noindent Let us even find $z_{n}$ such that $(2n+1)^{4V_{\mathcal{C}}/z_{n}%
}=\frac{\alpha^{-z_{n}n}}{1-\alpha}$. It is thus equivalent to: $-n\ln
(\alpha)z_{n}^{2}-\ln(1-\alpha)z_{n}-4V_{\mathcal{C}}\ln(2n+1)=0$

\noindent We have $\Delta=\ln(1-\alpha)^{2}-16V_{\mathcal{C}}\ln
(2n+1)n\ln(\alpha)>0$ since $|\alpha|<1$

\noindent Since $0<z_{n}<1$, we have necesseraly $z_{n}$ the non negative root
of the previous equation which leads to:
\begin{align*}
z_{n}  &  =\frac{\ln(1-\alpha)+\sqrt{\ln(1-\alpha)^{2}-16V_{\mathcal{C}}%
\ln(2n+1)n\ln(\alpha)}}{-2n\ln(\alpha)}\\
&  \sim\frac{4V_{\mathcal{C}}^{1/2}}{\ln(1/\alpha)^{1/2}}\sqrt{\frac{\ln
(n)}{n}}\\
&  \sim\frac{2\sqrt{2}V_{\mathcal{C}}^{1/2}}{\eta}\sqrt{\frac{\ln(n)}{n}}%
\end{align*}

\noindent We can inject $z_{n}$ in $(2n+1)^{4V_{\mathcal{C}}/z_{n}}\alpha
^{n}+\frac{\alpha^{(1-z_{n})n}}{1-\alpha}$ and we find that
\[
\Pr(\widetilde{R}_{n}(\phi_{n}^{B,p_{n}^{\star}})-\hat{R}_{CV}^{Out}%
(p_{n}^{\star})\geq\eta)=O_{n}((n+1)^{8V_{\mathcal{C}}}\exp(-2n(\eta-2\sqrt
{2}V_{\mathcal{C}}^{1/2}\sqrt{\ln(n)/n})^{2})/(1-\exp(-2\eta^{2}))
\]

\noindent Let us now find the expression of $f$ the inverse of $\min
_{\varepsilon}(B(n,p_{n},\varepsilon),V(n,p_{n},\varepsilon))$ with

\begin{itemize}
\item $B(n,p_{n},\varepsilon)=\displaystyle     \min((2n(1-p_{n}%
)+1)^{^{\frac{4V_{\mathcal{C}}}{1-p_{n}}}}\exp(-n\varepsilon^{2}/9))$

\item $V(n,p_{n},\varepsilon)=\displaystyle     \exp(-2np_{n}\varepsilon^{2}).$
\end{itemize}

\noindent In the case of ERM\ algorithm,
\[
\exp(-2np_{n}\varepsilon^{2})\leq(2n(1-p_{n})+1)^{^{\frac{4V_{\mathcal{C}}%
}{1-p_{n}}}}\exp(-n\varepsilon^{2}/9)
\]
if and only if $-2np_{n}\varepsilon^{2}\leq\frac{4V_{\mathcal{C}}}{1-p_{n}}%
\ln(2n(1-p_{n})+1)-n\varepsilon^{2}/9$ which is equivalent to%
\[
n(1/9-2p_{n})\varepsilon^{2}\leq\frac{4V_{\mathcal{C}}\ln(2n(1-p_{n}%
)+1)}{1-p_{n}}%
\]
and also $\varepsilon\leq\sqrt{\frac{4V_{\mathcal{C}}\ln(2n(1-p_{n}%
)+1)}{n(1-p_{n})(1/9-2p_{n})}}:=\varepsilon_{n}$.

\bigskip

\noindent Thus if $\varepsilon\leq\varepsilon_{n}$, it follows that
$\min(B(n,p_{n},\varepsilon),V(n,p_{n},\varepsilon))=\exp(-2np_{n}%
\varepsilon^{2})$, thus if $\delta=\exp(-2np_{n}\varepsilon^{2})$ we deduce
that $\varepsilon=\sqrt{\frac{\ln(1/\delta)}{np_{n}}}$. If $\varepsilon
>\varepsilon_{n}$, $\min(B(n,p_{n},\varepsilon),V(n,p_{n},\varepsilon
))=(2n(1-p_{n})+1)^{^{\frac{4V_{\mathcal{C}}}{1-p_{n}}}}\exp(-n\varepsilon
^{2}/9)$. Thus if $\delta=(2n(1-p_{n})+1)^{^{\frac{4V_{\mathcal{C}}}{1-p_{n}}%
}}\exp(-n\varepsilon^{2}/9)$, we then deduce that $\varepsilon=3\sqrt
{\frac{4V_{\mathcal{C}}\ln(2n(1-p_{n})+1)/(1-p_{n})+\ln(1/\delta)}{n}}$.
Denote $\delta_{n}=\exp(-2np_{n}\varepsilon_{n}^{2})=\exp(-\frac{4p_{n}%
V_{\mathcal{C}}\ln(2n(1-p_{n})+1)}{(1-p_{n})(1/9-2p_{n})})=(2n(1-p_{n}%
)+1)^{-\frac{4p_{n}V_{\mathcal{C}}}{(1-p_{n})(1/9-2p_{n})}}$.

\bigskip

\noindent In conclusion, if $\delta\geq\delta_{n}$, we have:
\[
f(n,p_{n},\delta)=\sqrt{\frac{\ln(1/\delta)}{2np_{n}}}%
\]
\noindent and if $\delta<\delta_{n}$,
\[
f(n,p_{n},\delta)=3\sqrt{\frac{4V_{\mathcal{C}}\ln(2n(1-p_{n})+1)/(1-p_{n}%
)+\ln(1/\delta)}{n}}\leq6\sqrt{\frac{V_{\mathcal{C}}\ln(2n+1)+\ln(1/\delta
)}{n(1-p_{n})}}.
\]

\bigskip

$\Box$

\noindent In summary, the probability of the deviation between the
out-of-bag cross-validation estimate and the generalization error
is bounded by the minimum of a Hoeffding-type bound and a
Vapnik-Chernovenkis-type bounds, and thus it is smaller than 1
even for small learning sets. Finally, we also give a simple rule
on how to subbag the predictor. However, in the case of
classification, we show that subagging strong learners can give a
strong learner. It would be more interesting to answer the
following question : can we obtain a similar result with the
subagging of weak learners ?

\newpage

\section{Appendices}

\noindent We will use the definition of strong difference bounded introduced
by \cite{KUT02} and a corollary of his main theorem inspired by \cite{McD89}.

\begin{definition}
[Kutin\cite{KUT02}]Let $\Omega_{1},\ldots,\Omega_{n}$ be probability spaces.
Let $\Omega=\prod_{k=1}^{n}\Omega_{k}$ and let $X$ a random variable on
$\Omega$. We say that $X$ is strongly difference bounded by $(b,c,\delta)$ if
the following holds: there is a \textquotedblright bad\textquotedblright
\ subset $B\subset\Omega$, where $\delta=\mathbb{P}(B)$. If $\omega
,\omega^{\prime}\in\Omega$ differ only in $k$-th coordinate, and $\omega\notin
B$, then%

\[
|X(\omega)-X(\omega^{\prime})| \leq c
\]

Furthermore, for any $\omega, \omega^{\prime}\in\Omega$,
\[
|X(\omega)-X(\omega^{\prime})| \leq b
\]
\end{definition}

\noindent We will need the following theorem. It says in substance that a
strongly difference bounded function of independent variables is closed to its
expectation with high probability.

\begin{theorem}
[Kutin\cite{KUT02}]\label{Kutin} Let $\Omega_{1},\ldots,\Omega_{n}$ be
probability spaces. Let $\Omega=\prod_{k=1}^{n}\Omega_{k}$ and let $X$ a
random variable on $\Omega$, which is strongly difference bounded by
$(b,c,\delta)$. Assume $b\geq c\geq0$ and $\alpha>0$. Let $\mu=\mathbb{E}(X)$.
Then, for any $\tau>0$,%

\[
\Pr(X-\mu\geq\tau)\leq2(\exp(-\frac{\tau^{2}}{8n(c+b\alpha)^{2}}%
)+\frac{n}{\alpha}\delta)
\]
\end{theorem}

\noindent We will use the definition of weak difference bounded introduced by
\cite{KUT02} and a corollary of his main theorem.

\begin{definition}
[Kutin]Let $\Omega_{1},\ldots,\Omega_{n}$ be probability spaces. Let
$\Omega=\prod_{k=1}^{n}\Omega_{k}$ and let $X$ a random variable on $\Omega$.
We say that $X$ is weakly difference bounded by $(b,c,\delta)$ if the
following holds: for any $k$,%

\[
\forall^{\delta}(\omega,v)\in\Omega\times\Omega_{k},\text{\ }\mathbb{P}%
(|X(\omega)-X(\omega^{^{\prime}})|)\leq c
\]
\noindent where $\omega_{k}^{^{\prime}}=v$ and $\omega_{i}^{^{\prime}}%
=\omega_{i}$ for $i\neq k$. and the notation $\forall^{\delta}\omega
,\Phi(\omega)$ means ''$\Phi(\omega)$ holds for all but but a $\delta$
fraction of $\Omega$''%

\[
|X(\omega)-X(\omega^{\prime})|\leq c
\]

\noindent Furthermore, for any $\omega,\omega^{\prime}\in\Omega$, differing
only one coordinate:
\[
|X(\omega)-X(\omega^{\prime})|\leq b
\]
\end{definition}

\noindent We will need the following theorem. It says in substance that a
weakly difference bounded function of independent variables is closed to its
expectation with probability.

\begin{theorem}
[Kutin]\label{KutinWeak}Let $\Omega_{1},\ldots,\Omega_{n}$ be probability
spaces. Let $\Omega=\prod_{k=1}^{n}\Omega_{k}$ and let $X$ a random variable
on $\Omega$.which is weakly difference bounded by $(b,c,\delta)$. Assume
$b\geq c\geq0$ and $\alpha>0$. Let $\mu=\mathbb{E}(X)$. Then, for any
$\varepsilon>0$%

\[
\Pr(|X-\mu|\geq\varepsilon)\leq2\exp(-\frac{\varepsilon^{2}}{10nc^{2}%
(1+\frac{2\varepsilon}{15nc})^{2}})+\frac{2nb\delta^{1/2}}{c}\exp
(\frac{\varepsilon b}{4nc^{2}}))+2n\delta^{1/2}%
\]
\end{theorem}


\clearpage

\begin{thebibliography}{99}
%
%
%
%
%
%
%
%
%
%
%
%
%
%
%
%
%
%\providecommand{\natexlab}[1]{#1} \providecommand{\url}[1]{\texttt{#1}}
%\expandafter\ifx\csname urlstyle\endcsname\relax \providecommand{\doi}[1]{doi: #1}\else %\providecommand{\doi}{doi: \begingroup\urlstyle{rm}\Url}\fi
%
%
%

\bibitem[AL68]{AL68}D. M. Allen, The relationship between variable selection
and data augmentation and a method for prediction. \textit{Technometrics}
1968, 16, 125-127.

\bibitem[ATE92]{ATE92}M. Atteia. Hilbertian kernels and spline functions.
North-Holland, 1992.

\bibitem[ARL07]{ARL07}S. Arlot, Model selection by resampling penalization.
\textit{submitted to COLT} 2007.

\bibitem[BEN04]{BEN04}Y. Bengio and Y. Grandvalet. No Unbiased Estimator of
the Variance of K-Fold Cross-Validation. \textit{Journal of Machine Learning
Research} 5, 1089-1105, 2004.

\bibitem[BIS05]{BIS05}M. Markatou, H. Tian, S. Biswas, G. Hripcsak. Analysis
of Variance of Cross-Validation Estimators of the Generalization Error.
\textit{Journal of Machine Learning Research} 1127-1168, 2005.

\bibitem[BREI84]{BREI84}L. Breiman, J. H. Friedman, R. Olshen, and C. J.
Stone. Classification and regression trees. \textit{The Wadsworth statistics
probability series}. Wadsworth International Group, 1984.

\bibitem[BREI92]{BREI92}L. Breiman, and Spector, P. (1992), Submodel selection
and evaluation in regression: The X-random case \textit{International
Statistical Review}, 60, 291-319.

\bibitem[BREI96]{BREI96}L. Breiman. Bagging predictors. \textit{Machine
Learning}, 24:123--140.

\bibitem[BKL99]{BKL99}A. Blum, A., Kalai, A., and Langford, J. (1999). Beating
the hold-out: Bounds for k-fold and progressive cross-validation.
\textit{Proceedings of the International Conference on Computational Learning
Theory}.

\bibitem[BE01]{BE01}O. Bousquet and A. Elisseef. Algorithmic stability and
generalization performance \textit{In Advances in Neural Information
Processing Systems} 13: Proc. NIPS'2000, 2001.

\bibitem[BE02]{BE02}O. Bousquet and A. Elisseef. Stability and generalization.
\textit{Journal of Machine Learning Research}, 2002.

\bibitem[BUR89]{BUR89}P. Burman. A comparative study of ordinary
cross-validation, v-fold cross-validation and the repeated learning-testing
methods. \textit{Biometrika}, 76:503-- 514, 1989.

\bibitem[BTW07]{BTW07}F. Bunea, A.B. Tsybakov and M.H. Wegkamp, M. H. Sparsity
oracle inequalities for the Lasso. \textit{Electron. J. Statist.}, 1 169?194, 2007.

\bibitem[COR09]{cornec}M.Cornec. Concentration inequalities of the
cross-validation estimator forEmpirical Risk Minimiser. Technical Report. 2009.

\bibitem[DGL96]{DGL96}L. Devroye, L. Gyorfi, and G. Lugosi. A Probabilistic
Theory of Pattern Recognition. Number 31 in \textit{Applications of
Mathematics}. Springer, 1996.

\bibitem[DW79]{DW79}L. Devroye and T. Wagner. Distribution-free performance
bounds for potential function rules. \textit{IEEE Trans. Inform. Theory},
25(5):601 604, 1979. 41

\bibitem[DEWA79]{DEWA79}L. P. Devroye and T. J. Wagner. Distribution-free
inequalities for the deleted and holdout error estimates. \textit{IEEE
Transactions on Information Theory}, IT?25(2):202?207, 1979

\bibitem[DUD03]{DUD03}S. Dudoit and M. J. van der Laan. Asymptotics of
cross-validated risk estimation in model selection and performance assessment.
\textit{Technical Report} 126, Division of Biostatistics, University of
California, Berkeley, 2003.

\bibitem[DUD04]{DUD04}S. Dudoit, M. J. van der Laan, S. Keles, A. M. Molinaro,
S. E. Sinisi, and S. L. Teng. Loss-based estimation with cross-validation:
Applications to microarray data analysis. \textit{SIGKDD Explorations,
Microarray Data Mining Special Issue}, 2004.

\bibitem[DUD04BIS]{DUD04BIS}M.J. van der Laan, S. Dudoit, A. van der Vaart
(2004),The cross-validated adaptive epsilon-net estimator, submitted for
publication in \textit{Statistics and Decisions}.

\bibitem[FRE95]{FRE95}Y. Freund and R. Schapire. A decision-theoretic
generalization of on-linelearning and an application to boosting. In
\textit{Proc. of the Second European Conference on Computational Learning
Theory}. LNCS, March 1995.

\bibitem[GEI75]{GEI75}S. Geisser. The predictive sample reuse method with
applications. \textit{Journal of the American Statistical Association},
70:320--328, 1975.

\bibitem[GYO02]{GYO02}L. Gyorfi, M. Kohler, A. Krzy?zak, and H. Walk.
\textit{A distribution-free theory of nonparametric regression}.
Springer-Verlag, New York, 2002a.

\bibitem[HTF01]{HTF01}T. Hastie, R. Tibshirani, and J. H. Friedman. The
Elements of Statistical Learning: Data Mining, Inference, and Prediction.
Springer-Verlag, 2001.

\bibitem[HOEF63]{HOEF63}W. Hoeffding, (1963). Probability inequalities for
sums of bounded random variables. \textit{Journal of the American Statistical
Association}, 58, 13?30.

\bibitem[HOL96]{HOL96}S. B. Holden. Cross-validation and the PAC learning
model. \textit{Research Note} RN/96/64, Dept. of CS, Univ. College, London, 1996.

\bibitem[HOL96bis]{HOL96bis}S. B. Holden. PAC-like upper bounds for the sample
complexity of leave-one-out cross validation. \textit{In Proceedings of the
Ninth Annual ACM Workshop on Computational Learning Theory}, pages 41 50, 1996.

\bibitem[KR99]{KR99}M. Kearns and D. Ron. Algorithmic stability and
sanity-check bounds for leave-one-out cross-validation. \textit{Neural
Computation}, 11:1427 1453, 1999.

\bibitem[KEA95]{KEA95}M. Kearns, (1995). A bound on the error of cross
validation, with consequences for the training-test split. \textit{In Advances
in Neural Information Processing Systems 8}. The MIT Press.

\bibitem[KMNR95]{KMNR95}M. J. Kearns, Y. Mansour, A. Ng,, and D. Ron. An
experimental and theoretical comparison of model selection methods. \textit{In
Proceedings of the Eighth Annual ACM Workshop on Computational Learning
Theory}, pages 21 30, 1995. To Appear in Machine Learning, COLT95 Special Issue.

\bibitem[KUT02]{KUT02}S. Kutin. Extensions to McDiarmid's inequality when
differences are bounded with high probability. \textit{Technical report},
Department of Computer Science, The University of Chicago, 2002. In preparation.

\bibitem[KUNIY02]{KUNIY02}S. Kutin and P. Niyogi.\ Almost-everywhere
algorithmic stability and generalization error, 2002. \textit{Technical
report} TR-2002-03, University of Chicago.

\bibitem[KUNIY01]{KUNIY01}S. Kutin and P. Niyogi. The interaction of stability
and weakness in AdaBoost. \textit{Technical Report} TR-2001-30, Department of
Computer Science, The University of Chicago, 2001.

\bibitem[LM68]{LM68}P. A. Lachenbruch,; M. Mickey, Estimation of error rates
in discriminant analysis. \textit{TechnometricsLM68} Estimation of error rates
in discriminant analysis. \textit{Technometrics} 1968, 10, 1-11.

\bibitem[Li87]{Li87}K-C Li. Asymptotic optimality for cp, cl, cross-validation
and generalized cross-validation: Discrete index set. \textit{Annals of
Statistics}, 15:958--975, 1987.

\bibitem[Lug03]{Lug03}G Lugosi. Concentration-of-measure inequalities
presented at \textit{the Machine Learning Summer School 2003}, Australian
National University, Canberra,

\bibitem[McC76]{McC76}P. J. McCarthy. The use of balanced half-sample
replication in crossvalidation studies. \textit{Journal of the American
Statistical Association}, 71: 596--604, 1976.

\bibitem[McD89]{McD89}C. McDiarmid. On the method of bounded differences.
\textit{In Surveys in combinatorics}, 1989 (Norwich, 1989), pages 148 188.
Cambridge Univ. Press, Cambridge, 1989.

\bibitem[McD98]{McD98}C. McDiarmid. Concentration. In Probabilistic Methods
for Algorithmic Discrete Mathematics, pages 195 248. Springer, Berlin, 1998.

\bibitem[PIC84]{PIC84}R. R. Picard and R. D. Cook. Cross-validation of
regression models. \textit{Journal of the American Statistical Association},
79:575--583, 1984.

\bibitem[RIP96]{RIP96}B. D. Ripley. Pattern recognition and neural networks.
\textit{Cambridge University Press}, Cambridge, New York, 1996.

\bibitem[ROG63]{ROG63}C. Rogers. Covering a sphere with spheres.
\textit{Mathematika}, vol. 10,pp. 157-164, 1963.

\bibitem[SHAO93]{SHAO93}J. Shao. Linear model selection by cross-validation.
\textit{Journal of the American Statistical Association}, 88:486--494, 1993. !

\bibitem[STO74]{STO74}M. Stone,(1974). Cross-validatory choice and assessment
of statistical predictions. \textit{Journal of the Royal Statistical Society
B}, 36, 111?147.

\bibitem[STO77]{STO77}M. Stone, (1977).Asymptotics for and against
cross-validation. \textit{Biometrika}, 64, 29?35.

\bibitem[VAL84)]{VAL84}L.G. Valiant (1984). A theory of learnable.
\textit{Proc. of the 1984, STOC}, pages 436-445.

\bibitem[Vaart96]{Vaart96}A. W. van der Vaart and J. Wellner. Weak Convergence
and Empirical \textit{Processes}. Springer-Verlag, New York, 1996.

\bibitem[VA71]{VA71}V. Vapnik, and A. Chervonenkis, (1971). On the uniform
convergence of relative frequencies of events to their probabilities.
\textit{Theory of Probability and its Applications}, 16, 264?280.

\bibitem[VC71]{VC71}V. N. Vapnik and A. Y. Chervonenkis. On the uniform
convergence of relative frequencies of events to their probabilities.
\textit{Theory of Probability and its Applications}, 16(2):264--280,1971.

\bibitem[VA82]{VA82}V. Vapnik, (1982). Estimation of Dependences Based on
Empirical Data. Springer-Verlag.

\bibitem[Vap95]{Vap95}V. Vapnik. The nature of statistical learning theory.
Springer, 1995.

\bibitem[Vap98]{Vap98}V. Vapnik. Statistical learning theory. John Wiley and
Sons Inc., New York, 1998. A Wiley-Interscience Publication.

\bibitem[YAN07]{YAN07}Y. Yang, Consistency of Cross Validation for Comparing
Regression Procedures. \textit{Accepted by Annals of Statistics}.

\bibitem[ZHA93]{ZHA93}P. Zhang. Model selection via multifold
cross-validation. \textit{Annals of Statistics}, 21:299--313, 1993.

\bibitem[ZHA00]{ZHA00}T Zhang . A leave-one-out cross validation bound for
kernel methods with applications in learning. \textit{14th Annual Conference
on Computational Learning Theory}, 2001 - Springer.
\end{thebibliography}
\end{document}